\documentclass{article}

\usepackage[final]{neurips_2021}

\usepackage[utf8]{inputenc} 
\usepackage[T1]{fontenc}    
\usepackage{hyperref}       
\usepackage{url}            
\usepackage{booktabs}       
\usepackage{amsfonts}       
\usepackage{nicefrac}       
\usepackage{microtype}      

\usepackage{threeparttable}
\usepackage{times}
\usepackage{epsfig}
\usepackage{graphicx}
\usepackage{amsmath}
\usepackage{amssymb}
\usepackage{xcolor}
\usepackage{pdfrender}
\newcommand*{\boldcheckmark}{%
	\textpdfrender{
		TextRenderingMode=FillStroke,
		LineWidth=.5pt, 
	}{\checkmark}%
}
\usepackage{cleveref}
\hypersetup{colorlinks}
\usepackage{mathtools}
\usepackage{algorithm}
\usepackage{dpr}
\usepackage{fec}
\usepackage{enumerate}
\usepackage{subcaption}
\usepackage[noend]{algpseudocode}
\usepackage{makecell}
\usepackage{comment}
\usepackage{caption}
\usepackage{subcaption}
\usepackage{enumitem}
\usepackage{amsthm}
\usepackage{arydshln}
\usepackage{bm}
\usepackage{siunitx,tabularx,ragged2e,booktabs}
\usepackage{wrapfig}
\usepackage{float}
\usepackage{booktabs}
\usepackage{pifont}
\usepackage{soul}
\newcommand{\xmark}{\ding{55}}%
\newcommand{\ilsvrc}{\text{ILSVRC2012}}
\newcommand{\imagenet}{\text{ImageNet}}

\newcommand{\rda}{\text{RDA}}
\newcommand{\proxsvrg}{\text{Prox-SVRG}}
\newcommand{\proxsg}{\text{Prox-SG}}
\newcommand{\proxspider}{\text{Prox-Spider}}
\newcommand{\saga}{\text{SAGA}}

\newcommand{\hspg}{\text{HSPG}}
\newcommand{\hspgname}{Half-Space Stochastic Projected Gradient}

\newcommand{\algacro}{OTO{}}

\newcommand{\cifar}{\text{CIFAR10}}
\newcommand{\fashionmnist}{\text{Fashion-MNIST}}
\newcommand{\resnet}{\text{ResNet18}}
\newcommand{\resnetfifty}{\text{ResNet50}}
\newcommand{\vgg}{\text{VGG16}}
\newcommand{\vggbn}{\text{VGG16-BN}}
\newcommand{\mobilenet}{\text{MobileNetV1}}
\newcommand{\bert}{\text{Bert}}
\newcommand{\squad}{\text{SQuAD}}

\newcommand{\halfspacestep}{\text{Half-Space Step}}

\newcommand{\zz}[1]{{\color{purple} #1}}
\newcommand{\ie}{\textit{i.e.}}
\newcommand{\eg}{\textit{e.g.}}

\newtheorem{theorem}{Theorem}
\newtheorem{lemma}{Lemma}
\newtheorem{corollary}{Corollary}

\newtheorem{definition}{Definition}
\newtheorem{assumption}{Assumption}
\newtheorem{proposition}{Proposition}

\def \B {\mathcal{B}}

\def \R {\mathbb{R}}

\def \M {\mathcal{M}}

\def \I {\mathcal{I}}
\def \S {\mathcal{S}}
\def \P {\mathcal{P}}

\def \G {\mathcal{G}}
\def \O {\mathcal{O}}
\def \K {\mathcal{K}}

\title{Only Train Once: A One-Shot Neural Network Training And Pruning Framework}

\author{%
	Tianyi Chen\thanks{Corresponding author.} \\
	Microsoft\\
	\texttt{tiachen@microsoft.com} \\
	\And
	Bo Ji\\
	National University of Singapore\\
	\texttt{jibo@comp.nus.edu.sg} \\
	\And
	Tianyu Ding \\
	Johns Hopkins University \\
	\texttt{tding1@jhu.edu} \\
	\And
	Biyi Fang \\
	Microsoft \\
	\texttt{bif@microsoft.com} \\
	\And
	Guanyi Wang \\
	Georgia Institute of Technology \\
	\texttt{gwang93@gatech.edu} \\
	\And
	Zhihui Zhu\\
	University of Denver\\
	\texttt{zhihui.zhu@du.edu}\\
	\And
	Luming Liang\\
	Microsoft\\
	\texttt{lulian@microsoft.com} \\  
	\And
	Yixin Shi\\
	Microsoft\\
	\texttt{yixshi@microsoft.com} \\  
	\And
	Sheng Yi\\
	Microsoft\\
	\texttt{shengyi@microsoft.com} \\  
	\And
	Xiao Tu\\
	Microsoft\\
	\texttt{xiaotu@microsoft.com} \\  
}

\begin{document}
	
	\maketitle
	
	\begin{abstract}
		
		Structured pruning is a commonly used technique in deploying deep neural networks (DNNs) onto resource-constrained devices. However, the existing pruning methods are usually heuristic, task-specified, and require an extra fine-tuning procedure. To overcome these limitations, we propose a framework that compresses DNNs into slimmer architectures with competitive performances and significant FLOPs reductions by Only-Train-Once (OTO).
		OTO contains two keys: $(i)$ 
		we partition the parameters of DNNs into zero-invariant groups, enabling us to prune zero groups without affecting the output; 
		and $(ii)$ to promote zero groups, 
		we then formulate a structured-sparsity optimization problem and propose a novel optimization algorithm,~\hspgname{}~(\hspg{}), to solve it, which outperforms the standard proximal methods on group sparsity exploration and maintains comparable convergence. 
		To demonstrate the effectiveness of \algacro{}, we 
		train and compress full models simultaneously from scratch without fine-tuning for inference speedup and parameter reduction, and achieve state-of-the-art results
		on VGG16 for CIFAR10, ResNet50 for CIFAR10 and Bert for SQuAD and competitive result on ResNet50 for ImageNet. The source code is available at~\url{https://github.com/tianyic/only_train_once}.

	\end{abstract}

	
	\section{Introduction}
	
	Deep neural networks (DNNs) have been shown to be effective in various real applications \cite{lecun2015deep,goodfellow2016deep}. It is widely acknowledged
	that large-scale DNN models not only learn faster but also outperform their slimmer counterparts. However, such heavy models pose a great challenge to the deployment stage due to their resource-consuming nature. In addressing this issue, many model compression techniques~\cite{bucilu2006model,cheng2017survey} are proposed in the past decade that aim at compressing those large and complex models into slimmer and simpler ones while suffering negligible loss in performance.

	Pruning methods
	as one of the main categories of model compression, focus on 
	identifying and pruning redundant structures via various mechanisms
	to achieve a slimmer architecture, and thus improve the interpretability of a DNN model~\cite{gale2019state,cheng2017survey,neklyudov2017structured}. For example,~\cite{han2015deep,han2015learning} adopt fine-grained pruning via $\ell_1$ or $\ell_2$ regularization, which prune the small-weight connections based on some hard threshold.~\cite{he2018soft,li2019exploiting,luo2017thinet} measure the importance of filters  
	to accelerate the networks by removing insignificant feature maps.~\cite{he2018amc,chen2019storage} utilize reinforcement learning agent to predict compression action. 
	
	
	Nevertheless, many of the existing pruning methods $(i)$ often rely on criteria  based on heuristics or empirical cues, \eg, magnitude of a connection weight and importance score of a filter, to identify redundant parameters, which may cause divergence during optimization;
	$(ii)$ thus require complex multi-stage pipelines that involve either a retraining or fine-tuning procedure to regain the accuracy during constructing a slimmer model, which is time-consuming;
	and $(iii)$ are specific to certain architectures or applications, and are consequently less applicable to various downstream scenarios. Recently, there have been a few efforts \cite{deleu2021structured,lin2019toward,chen2020orthant} to directly train the network with sparsity inducing regularizers, which provide generality and convergence guarantee. However, these approaches focus on either merely the individual sparsity of the parameters or the group sparsity of the filters, and thus cannot directly remove those zero components (still require subsequent fine-tuning) since the zeros are entangled with other commonly used components, \eg, bias, batch normalization or skip connection. Furthermore, the optimization algorithms used in~\cite{deleu2021structured,lin2019toward} lack sufficient capability to explore (group) sparsity in DNNs effectively and require a post-processing step to yield exact zeros.

	\begin{figure*}[t]
		\hspace{-2mm}
		\includegraphics[scale=0.652]{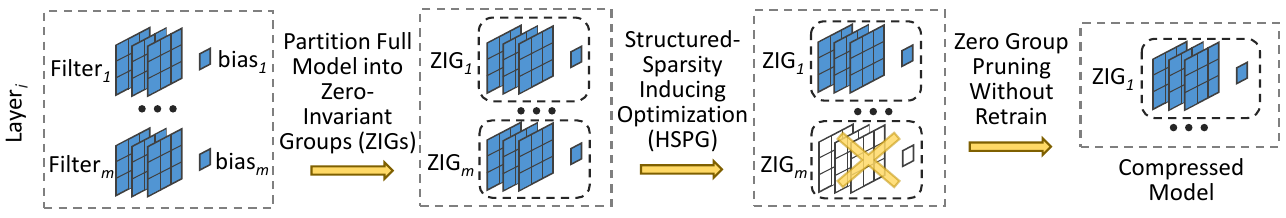}
		\vspace{-4mm}
		\caption{Overview of \algacro{}. Without loss of generality, we illustrate \algacro{} on a model with only vanilla convolutional layers, and for simplicity we only show Layer$_i$ with $m$ 3D filters and their biases. The key to its success is twofold: $(i)$ identify and partition the parameters of the model into zero-invariant groups (ZIGs); and $(ii)$ solve the structured-sparsity regularization problem using \hspg{}. Finally, we obtain the compressed model by directly pruning the zero groups, \ie,~ZIG$_m$.}
		\vspace{-5mm}
		\label{fig.overview}
	\end{figure*}

	In this paper, we overcome the above limitations of existing pruning methods by proposing a \emph{one-shot} neural network pruning framework, with which we are able to train a full heavy model from scratch only once, and obtain a slim architecture without fine-tuning while maintain high performance. As shown in Figure~\ref{fig.overview}, the key to its success is twofold: $(i)$ we identify and partition the parameters of DNNs into zero-invariant groups (ZIGs), enabling us to prune redundant structures according to zero groups without affecting the output of the network; and $(ii)$ to promote zero groups, we formulate the pruning task as a structured-sparsity optimization problem and propose a novel optimization method,~\hspgname{}~(\hspg{}), to solve it, which outperforms the standard proximal methods on sparsity exploration and maintains comparable convergence. 
	We highlight that both zero-invariant group partition and the novel optimization algorithm in promoting zero group lead to achieve one-shot neural network training and pruning regardless of its architecture. 
	
	Our main contributions are summarized as follows.

	\vspace{-2mm}
	\begin{itemize}[leftmargin=*]
		\item \textbf{One-Shot Training and Pruning.} We propose \algacro{}, a training and pruning framework that compresses a full neural network with competitive performance
		by Only-Train-Once, thereby one-shot.
		\algacro{} dramatically simplifies the complex multi-stage training pipelines of the existing pruning approaches, 
		fits various architectures and applications, 
		and hence is generic and efficient.
		%

		
		\vspace{-1mm}
		\item \textbf{Zero-Invariant Group.} We define zero-invariant groups for neural networks.
		If a network is partitioned into ZIGs, it allows us to prune the zero groups without affecting the output, which results in one-shot pruning.
		Such property is applicable to various popular structures from plain fully connected layers to sophisticated ones such as residual blocks and multi-head attention.

		\vspace{-1mm}
		\item \textbf{Novel Structured-Sparsity Optimization Algorithm.}  
		We propose Half-Space Stochastic Projected Gradient (HSPG), a method that solves structured-sparsity inducing regularization problem.
		We show and analyze the superiority of \hspg{} in promoting zero groups of networks than the standard proximal methods and the competitive objective convergence in practice. 
		%
		The fact that ZIG and HSPG are designed agnostic to networks makes \algacro{} generic to various applications. 
		
		\vspace{-1mm}
		\item \textbf{Experimental Results.} We train and compress full models simultaneously from scratch without fine-tuning for inference speedup and parameter reduction, and achieve state-of-the-art results on compression benchmark VGG for CIFAR10, ResNet50 for~\cifar{}/ImageNet, Bert for SQuAD. 

	\end{itemize}

	\section{Related Work}

	Structured pruning focuses on identifying and pruning the redundant structures in a full model to achieve slimmer architectures for efficient model inference and storage~\cite{gale2019state, han2015deep}, where there have been numerous efforts dedicated. For CNN compression, the general procedure can be largely summarized as: (\textit{i}) train a full model; (\textit{ii}) identify and prune the redundant structures to build a slimmer model based on various criteria, including (structured) sparsity~\cite{lin2019toward,wen2016learning,deleu2021structured,li2020group,zhuang2020neuron,gao2020highly,zhuang2020neuron,meng2020pruning,yang2019deephoyer}, Bayesian pruning~\cite{zhou2019accelerate,neklyudov2017structured,louizos2017bayesian,van2020bayesian}, ranking importance~\cite{li2020eagleeye,luo2017thinet,hu2016network,he2018soft,li2019exploiting,zhang2018systematic}, reinforcement learning~\cite{he2018amc,chen2019storage}, adversarial robustness~\cite{sehwag2020hydra}, scientific control~\cite{tang2020scop},~lottery ticket~\cite{frankle2018lottery,frankle2019stabilizing,renda2020comparing}, joint quantization learning~\cite{Tung_2018_CVPR,yang2020automatic}, etc.; (\textit{iii}) retrain or iteratively fine-tune the slimmer model to regain the accuracy regression during pruning. 
	%
	These methods cannot avoid the extra and usually time-consuming fine-tuning step because the identified redundant structures, even parametrized with zeros, actually contribute to the model output, thereby additional fine-tuning step is an absolute necessity.
	%
	%
	
	For pruning \bert{}~\cite{NIPS2017_3f5ee243}, knowledge distillation~\cite{hinton2015distilling} and LayerDropout~\cite{fan2019reducing} shorten Bert by reducing the number of layers directly. 
	Other methods~\cite{gordon2020compressing,sanh2020movement,guo2019reweighted} build slimmer Berts in the manner of individual sparsity, but require specially designed data structure for storage and computing library to take advantage of sparse data~\cite{han2016eie,chen2018escoin}, and typically cannot achieve inference speedup against the highly optimized library~\cite{onnxruntime} for dense model due to the discontiguous memory allocation~\cite{chen2020spatially}. 
	
	The structured sparsity for weight pruning is the most relevant to the algorithm described in this paper. The existing structure learning works~\cite{lin2019toward,wen2016learning,deleu2021structured,li2020group,zhuang2020neuron} have the respective disadvantages: \textit{(i)} multiple trainings during the whole procedure since their group partition cannot isolate the impact of pruned structures to the model output; and \textit{(ii)} heuristic post-processing to generate zero groups as the standard proximal methods~\cite{duchi2009efficient,xiao2010dual,xiao2014proximal,defazio2014saga} and ADMM~\cite{zhang2018systematic,lin2019toward,boyd2011distributed} defective on the sparsity exploration for deep learning~\cite{chen2020orthant}, which may deteriorate the performance of the model significantly. 
	
	Avoiding fine-tuning step during the whole pruning procedure is receiving more and more attentions because of its efficiency. In particular, SNIP~\cite{lee2018snip} and GraSP~\cite{wang2020picking} identify redundancy via salience scores at the initialization stage to construct pruned structures, then train the pruned models by the standard optimizers.  SCP~\cite{kang2020operation} isolates the impact of batch normalization, while lacks the consideration of more general DNN architectures.

	\section{\algacro{}}
	%
	%
	In essence, \algacro{} frames the network training and pruning as a structure learning problem.
	Given a full model $\M$, \algacro{} trains and compresses it simultaneously from scratch \emph{without} fine-tuning, and achieves significant reduction in both FLOPs and parameters. 
	Particularly, as stated in Algorithm~\ref{alg:main.outline}, the trainable parameters of $\M$ are firstly partitioned into a ZIG set $\G$ (Section~\ref{sec.zig}). 
	We then construct and solve a structured-sparsity inducing optimization problem (Section~\ref{sec.formulation}) by proposed stochastic optimizer~(\hspg{}) to seek a highly group-sparse solution $\bm{x}^*_{\text{HSPG}}$ (Section~\ref{sec.hspg}).
	%
	Lastly, we obtain a compressed model $\M^*$ by directly pruning these zero groups
	(Section~\ref{sec.prune}).

	\vspace{-1mm}
	\begin{algorithm}[h!]
		\caption{Outline of \algacro{}.}
		\label{alg:main.outline}
		\begin{algorithmic}[1]
			\State \textbf{Input:} Full model $\M$ (no need to be pretrained). 
			\State \textbf{Construct $\bm\G$:} Partition the trainable parameters of $\mathcal{M}$ into a ZIG set $\G$.
			\State \textbf{Train:} Train the model $\mathcal{M}$ using~\hspg{}~(Algorithm.~\ref{alg:main.hspg.outline}) 
			to obtain a group-sparse solution $\bm{x}^*_{\text{HSPG}}$.
			\State \textbf{Prune:} Construct a slimmer model architecture $\M^*$ by directly pruning zero groups of $\bm{x}^*_{\text{HSPG}}$.
			\State \textbf{Output:} Compressed model $\M^*$.
		\end{algorithmic}
	\end{algorithm}
	\vspace{-3mm}
	
	\subsection{Zero-Invariant Group}\label{sec.zig}
	
	The root cause of the existing methods having multi-stage training pipeline is that despite the pruned structure (\eg,~3D filter) being zeros, its associated structures (\eg,~non-zero bias) still contribute to its corresponding output to the next layer (\eg,~feature map). 
	As a result, the model accuracy regresses, hence an extra step of fine-tuning is necessary.
	\algacro{} avoids the necessity by partitioning the parameters of DNNs into a set of so-called zero-invariant groups (ZIGs) $\mathcal{G}$ defined as follows.
	\begin{definition}[Zero-Invariant Groups (ZIGs)]
		\label{def:zero_invariant}
		For a layer-wise DNN, we partition its entire trainable parameters into disjoint groups $\G = \{g\}$. Then we call $\G$ zero-invariant groups (ZIGs) if each group $g\in \G$ is zero-invariant in the sense that all of the parameters in $g$ being zeros results in its corresponding output to the next layer to be zeros as well.
	\end{definition}
	\vspace{-2mm}
	
	In effect, if and only if a DNN model is partitioned into a ZIG set $\G$ and one or more of its element $g$ are parameterized by zeros, the entire corresponding structures contribute none to the model outputs and hence can be pruned directly.
	Such partition is applicable to various structures of DNN models.
	Without loss of generality, we define and describe ZIG partition for three most popular structures: \textit{(i)} Conv-BN, \textit{(ii)} Residual Block, and \textit{(iii)} Fully Connected and Multi-Head Attention Layer.
	
	\begin{figure}[htp!]
		\centering
		\begin{subfigure}{\linewidth}
			\centering
			\includegraphics[width=0.95\linewidth]{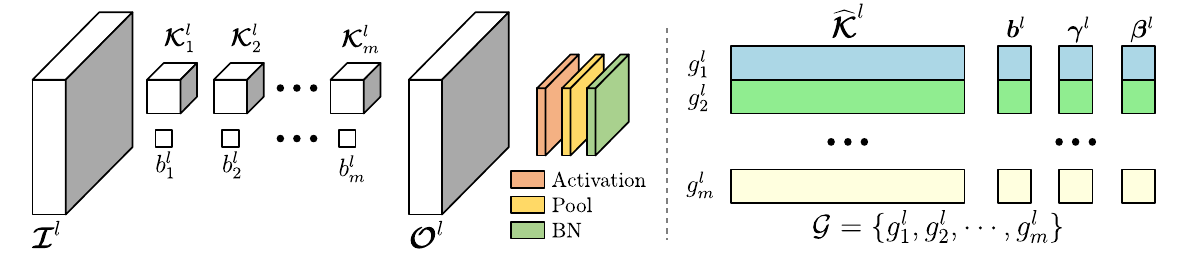}
			\vspace{-1.5mm}
			\caption{Conv-BN. $m$ denotes the number of channels in $\bm{\mathcal{O}}^{l}$.}
			\label{fig:convlayer_zig}
		\end{subfigure}
		\begin{subfigure}{\linewidth}
			\centering
			\includegraphics[width=0.95\linewidth]{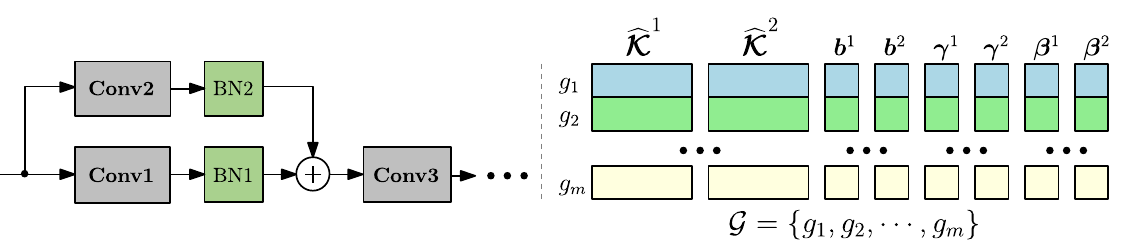}
			\vspace{-2.5mm}
			\caption{Residual block. $m$ denotes the number of output channels of the residual block.}\label{fig:residual_block_zig}
			\label{fig:1a}		
		\end{subfigure}
		\begin{subfigure}{\linewidth}
			\centering
			\includegraphics[width=0.75\linewidth]{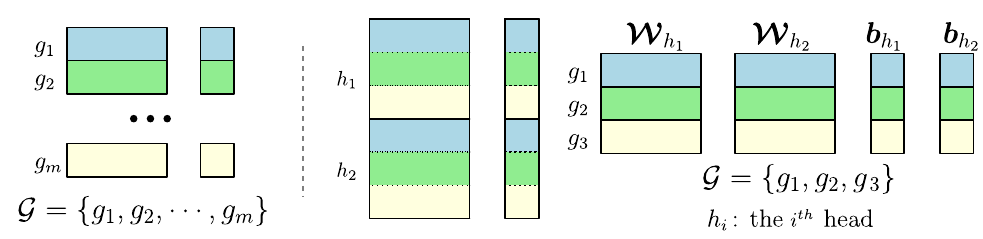}
			\caption{Fully connected layer (Left). Multi-head attention layer (Right). $m$ denotes the length of output vector.}\label{fig:fc_layer_zig}		
		\end{subfigure}
		\caption{Zero-invariant group partition for three popular structures.}
		\vspace{-0.1in}
		\label{fig.zip_group_partition}
	\end{figure}

	\textbf{ZIG of Conv-BN.}
	Convolutional layer (Conv) followed by batch-normalization layer (BN) is extensively used in DNN models. 
	Figure~\ref{fig:convlayer_zig} shows the ZIG partition for Conv-BN. 
	The 4D filter tensor {\scriptsize $\bm{\K}^{l}$} is flattened into a filter matrix {\scriptsize $\bm{\hat{\K}}^{l}$}.
	During the forward pass, the input tensor {\small $\bm{\I}^{l}$} is transformed into the output tensor {\small $\bm{\mathcal{O}}^{l}$} of Conv and then into the input tensor of the $(l+1)^{th}$ layer {\small $\bm{\mathcal{I}}^{l+1}$} by
	\vspace{-1mm}
	\begin{equation}\label{eq:convolution_operation}
		\bm{\mathcal{O}}^{l}\gets \bm{\mathcal{I}}^l \otimes \bm{\hat{\K}}^{l} +\bm{b}^l,\  \bm{\mathcal{I}}^{l+1}\gets\frac{a(\bm{\mathcal{O}}^{l})-\bm{\mu}^l}{\bm{\sigma}^{l}}\odot\bm{\gamma}^l +\bm{\beta}^l,
	\end{equation}
	
	where denoted by $\otimes$ the convolutional operation, $\odot$ the element-wise multiplication and $a(\cdot)$ the activation function. BN is parameterized by mean $\bm{\mu}^{l}$, standard deviation $\bm{\sigma}^l$, weight $\bm{\gamma}^l$ and bias $\bm{\beta}^l$ respectively.
	The activation function needs to be zero-invariant,~\ie, $a(\bm{0})=\bm{0}$, where most instances satisfy,~\eg, ReLU~\cite{fukushima1969visual}, PReLU~\cite{he2015delving}, GELU~\cite{hendrycks2016gaussian} and LeakyReLU \cite{xu2015empirical}. 
	Hence, each row of the flattened filter matrix {\scriptsize $\bm{\hat{\K}}^{l}$} and its bias $\bm{b}^l$ belong to one ZIG because they being zeros results in their corresponding channel of {\small $\bm{\mathcal{O}}^l$} (\ie, feature map) to be zeros as well.
	Subsequently, $\bm{\gamma}^l$ and $\bm{\beta}^l$ of this corresponding channel in BN are also included into this ZIG to avoid the value shift (zero to non-zero) during normalization.
	Note that grouping these four sets of parameters channel-wisely makes Conv-BN zero-invariant regardless of the value of $\bm{\mu}^{l}$ and $\bm{\sigma}^l$, and hence they are excluded from the ZIG.
	For illustration, each ZIG is highlighted in the same color (\eg, $g^{l}_{1}$ in blue).

	\textbf{ZIG of Residual Block.} 
	The residual block adds another layer of challenge because its output tensor is the summation of the outputs of two Conv-BNs. 
	Figure~\ref{fig:residual_block_zig} shows the ZIG partition for the residual block. 
	As illustrated, before propagated to Conv3, the outputs of Conv1-BN1 and Conv2-BN2 are summarized and hence share the same dimension. 
	As such, to make residual block zero-invariant, we group the four sets of parameters channel-wisely of both Conv1-BN1 and Conv2-BN2 into ZIGs, \ie, each row of {\scriptsize $\bm{\hat{\K}}^{1}$}, $\bm{b}^1$, $\bm{\gamma}^1$, $\bm{\beta}^1$ of Conv1-BN1 and each row of {\scriptsize $\bm{\hat{\K}}^{2}$}, $\bm{b}^2$, $\bm{\gamma}^2$, $\bm{\beta}^2$ of Conv2-BN2.
	In Appendix~\ref{appendix.implementation.zigresnet}, we describe the zero-invariant group partition of \resnetfifty{} in greater detail.
	
	\textbf{ZIG of Fully Connected and Multi-Head Attention Layer.}
	Figure~\ref{fig:fc_layer_zig} shows the ZIG partition for fully connected and multi-head attention layer. 
	Particularly, we partition each row of weight matrix and its associated bias into a ZIG, and therefore any input element is turned to zero if that ZIG is parameterized with zeros, making the fully connected layer zero-invariant.
	Multi-head attention layer is the key building block of the transformer architectures~\cite{NIPS2017_3f5ee243}. 
	Its trainable parameters contain a weight matrix and bias vector, consisting of the sub-matrix and sub-vector of each head (we use two heads as an example).
	We form ZIG by grouping each row of every sub-matrix and sub-vector, \ie, each row of {\footnotesize $\bm{\mathcal{W}_{h_{1}}}$}, $\bm{b_{h_{1}}}$,  {\footnotesize $\bm{\mathcal{W}_{h_{2}}}$} and $\bm{b_{h_{2}}}$ of $h_{1}$ and $h_{2}$, respectively.
	
	\textbf{Automatic ZIG Partition.} Based on the above illustrating examples, we provide prescribed ZIG partition for the tested DNNs in Section~\ref{sec.experiment}. 
	Furthermore, given an arbitrary DNN architecture, the procedure of partitioning variables into ZIGs  could be automatically proceeded, wherein the key would be identifying the connections among various layers, then performing corresponding group partition. We will leave the automatic ZIG partition for arbitrary DNNs as future work.

	\subsection{Structured-Sparsity Regularization} \label{sec.formulation}
	
	We now formulate a structured-sparsity regularization problem over the ZIG set $\G$ for the trainable parameters of the full model $\M$ as follows
	\begin{equation}\label{prob.main}
		\minimize{\bm{x}\in\mathbb{R}^n}\ \psi(\bm{x}):=f(\bm{x})+\lambda r(\bm{x}), \ r(\bm{x}):= \sum_{g\in\mathcal{G}}\norm{[\bm{x}]_g},
	\end{equation}
	where $\lambda>0$ is a weighting coefficient, $f(\bm{x})$ is a task-specific loss function,
	and $r(\bm{x})$ is an augmented structured-sparsity inducing regularization term 
	encoding the topological structure of $\M$ over $\G$. A larger $\lambda$ typically results in a higher group sparsity while sacrifices more on the bias of model estimation.
	We aim at computing a local optimum 
	to achieve both low loss and high group sparsity.

	To induce group sparsity onto the solution of~\eqref{prob.main}, there exist several candidates for $r(\bm{x})$, including mixed $\ell_1/\ell_p$ norm $(p>1)$~\cite{bach2012structured,el2018combinatorial} and group Minmax Concave Penalty (MCP)~\cite{zhang2010nearly}. Among these candidates, the mixed $\ell_1/\ell_2$ norm as defined in~\eqref{prob.main} is arguably the most popular choice in classical machine learning applications~\cite{bach2012structured,yang2010online},
	where $\norm{\cdot}$ is the $\ell_2$-norm, and each component $g\in\mathcal{G}$ indexes a group of variables. In this paper, we will demonstrate the effectiveness of~\algacro{} by selecting $r(\bm{x})$ as the mixed $\ell_1/\ell_2$ norm. We highlight~\algacro{} is applicable for other group sparsity regularizers as well.

	\subsection{\hspgname{}~(\hspg{})}\label{sec.hspg}
	
	To solve the non-smooth regularization problem as~\eqref{prob.main} in deep learning applications, the standard proximal method and the ADMM lack capability to effectively identify group sparsity; see the discussions later in this Section.
	Therefore, we propose a novel stochastic optimization algorithm so-called~\hspgname{}~(\hspg{}) to enhance the group sparsity exploration more effectively than the classical methods while maintain a similar convergence property.
	
	\textbf{Outline.} We state the outline of~\hspg{} in Algorithm~\ref{alg:main.hspg.outline}. It contains two stages: Initialization Stage and Group-Sparsity Stage. The first Initialization Stage employs Stochastic Gradient Descent (SGD) step to search for a good but usually non-sparse solution estimate. Then the second stage proceeds Half-Space step 
	started with the non-sparse iterate to effectively exploit the group sparsity within a sequence of reduced spaces and converges to the group-sparse solutions. Half-Space step performs SGD update on free non-zero variables along with a novel projection operator so-called Half-Space Projection, which significantly outperforms the standard proximal operators on sparsity exploration.

	\textbf{Initialization Stage.} The Initialization Stage performs the vanilla SGD to find a good initial point for the subsequent Group-Sparsity Stage.
	At $k^{th}$ iteration, a stochastic gradient of $f$, \eg, based on a mini-batch, is generated denoted as $\nabla \tilde{f}$. 
	Since the group sparsity inducing regularizer $r(\bm{x})$ in the form as~\eqref{prob.main} is non-smooth, we select a subgradient $\zeta(\bm{x}_k)$ from its subdifferential $\partial r(\bm{x}_k)$ to form a stochastic subgradient of $\psi(\bm{x}_k)$ as $\nu(\bm{x}_k):=\nabla \tilde{f}(\bm{x}_k)+\lambda \zeta(\bm{x}_k)$. We then compute the next iterate as  $\bm{x}_{k+1}:=\bm{x}_k-\alpha_k\nu(\bm{x}_k)$ by subgradient descent update.

	\begin{figure*}[ht]
		\centering
		\begin{subfigure}{0.48\textwidth}
			\centering
			\includegraphics[scale=0.56]{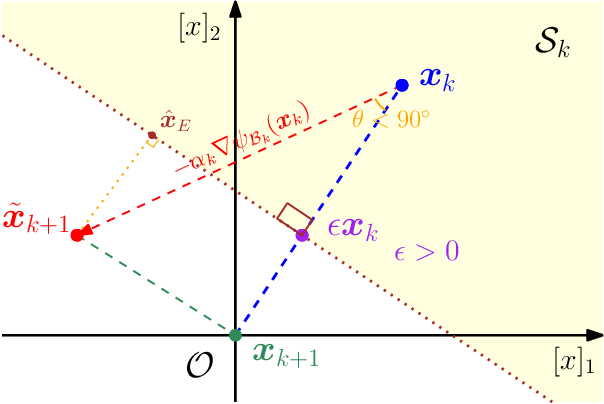}
			\caption{Half-Space Projection}
			\label{figure:half_space_projection}
		\end{subfigure}
		\begin{subfigure}{0.48\textwidth}
			\centering
			\includegraphics[scale=0.60]{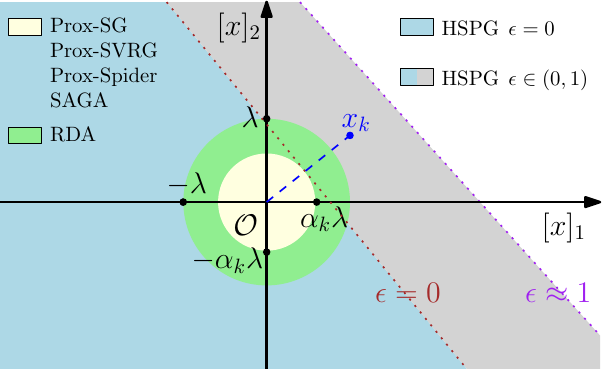}
			\caption{Projection Region For Mixed $\ell_1/\ell_2$ Regularization}
			\label{figure:project_region}
		\end{subfigure}
		\caption{Illustration of Half-Space Step with projection in \eqref{def:proj}, where $\mathcal{G}=\{\{1,2\}\}$.}
		\label{figure:proj_euclidean}
		\vspace{-0.3cm}
	\end{figure*}
	
	\textbf{Group-Sparsity Stage.} The Group-Sparsity Stage is designed to effectively determine the groups of zero variables and capitalize convergence characteristic,
	which is in sharp contrast to other heuristic aggressive weight pruning methods that typically lack theoretical guarantees~\cite{li2016pruning,luo2017thinet}. The intuition of~\halfspacestep{} is to project $[\bm{x}_k]_g$ to zero only if $-[\bm{x}_k]_g$ serves as a descent 
	step to $\psi(\bm{x}_k)$, \ie, $-[\bm{x}_k]_g^\top[\Grad \psi(\bm{x}_k))]_g<0$, hence updating $[\bm{x}_{k+1}]_g\gets[\bm{x}_k]_g-[\bm{x}_k]_g=0$ still results in some progress to the optimality. In particular, we first define the following index sets for any $ \bm{x}\in\mathbb{R}^n $:
	\begin{equation}\label{def:I_set}
		\mathcal{I}^0(\bm{x}) := \{g: g\in\mathcal{G}, [\bm{x}]_g=0\}\ \ \text{and}\ \ \mathcal{I}^{\neq 0}(\bm{x}) :=\{g: g\in\mathcal{G}, [\bm{x}]_g\neq 0\},
	\end{equation}
	where $\I^{0}(\bm{x})$ represents the indices of groups of zero variables at $\bm{x}$, and $\I^{\neq 0}(\bm{x})$ indexes the  groups of nonzero variables at $\bm{x}$. To proceed, we further define an artificial set that $\bm{x}$ lies in: 
	\begin{equation}\label{def:polytope}
		\S(\bm{x})\coloneqq \{\bm{0}\} \bigcup \{\bm{z}\in\mathbb{R}^n : [\bm{z}]_g=\bm{0} \ \text{if}\ g\in\I^0{(\bm{x})},\text{and}\  [\bm{z}]_g^\top[\bm{x}]_g\geq \epsilon \norm{[\bm{x}]_g}^2\ \text{if}\ g\in\I^{\neq 0}(\bm{x})\} ,
	\end{equation}
	which consists of half-spaces and the origin. Here the parameter $\epsilon \geq 0$ controls how aggressively we promote group sparsity,
	and is typically fixed as zero in practice.
	Hence, $\bm{x}\in\S_k:=\S(\bm{x}_k)$ only if: \textit{(i)} $[\bm{x}]_g$ lies in the upper half-space for all $g\in\mathcal{I}^{\neq 0}(\bm{x}_k)$ for some prescribed $\epsilon\in[0,1)$ as shown in Figure~\ref{figure:half_space_projection}; and \textit{(ii)} $[\bm{x}]_g$ equals to zero for all $g\in\mathcal{I}^0(\bm{x}_k)$. Intuitively, $\S_k$ establishes the region where important structures inhabit, thereby redundant structures vanish if falling outside.

	\begin{wrapfigure}{r}{0.6\textwidth}
		\vspace{-7mm}
		\begin{minipage}{\linewidth}
			\begin{algorithm}[H]
				\caption{Outline of \hspg{} for solving \eqref{prob.main}.}
				\label{alg:main.hspg.outline}
				\begin{algorithmic}[1]
					\State \textbf{Input:} $\bm{x}_0\in\mathbb{R}^n$, $ \alpha_0>0, \epsilon\in [0,1)$, and $ N\in \mathbb{Z}^+ $.
					\State \textbf{Output:} a group-sparse solution $\bm{x}^*_{\text{HSPG}}$ from $\{\bm{x}_{k}\}$.  
					\For{$k = 0,1,2,\dots$ } 
					\State Compute a stochastic subgradient $\nu(\bm{x}_k)$
					of $\psi(\bm{x}_k)$.\label{line:compute_subgrad_psi}
					\If{$k< N$} \label{line:switch_prox_sg_step}
					\State \textbf{\textit{Subgradient Descent Update:}}
					\State Set $\bm{x}_{k+1}\leftarrow\bm{x}_k-\alpha_k\nu(\bm{x}_k)$. 
					\Else{}
					\State \textbf{\textit{Half-Space Update:}}\label{line:half_space_start}
					\State Set a trial iterate $\tilde{\bm{x}}_{k+1}$ as 
					\vspace{-1mm}
					\begin{align*}
						[\tilde{\bm{x}}_{k+1}]_{\mathcal{I}^{\neq 0}(\bm{x}_k)}&\gets [\bm{x}_{k}-\alpha_k\nu(\bm{x}_k)]_{\mathcal{I}^{\neq 0}(\bm{x}_k)}\\
						[\tilde{\bm{x}}_{k+1}]_{\mathcal{I}^{ 0}(\bm{x}_k)}&\gets \bm{0}.
					\end{align*}
					\vspace{-5mm}
					\label{line:half_space_trial_iterate}
					\For{each group $ g $ in $\mathcal{G}$
					}\label{line:half_space_project_start}
					\State $[\bm{x}_{k+1}]_g\gets [\proj_{\mathcal{S}_k}^{HS}(\tilde{\bm{x}}_{k+1})]_g.$\label{line:half_space_end}
					\EndFor
					\EndIf
					\State Update $\alpha_{k+1}$. 
					\EndFor
				\end{algorithmic}
			\end{algorithm}
		\end{minipage}
		\vspace{-1mm}
	\end{wrapfigure}
	
	Ideally, the~Initialization Stage has produced reasonably well but typically non-sparse iterate $\bm{x}_k$ nearby a group-sparse solution $\bm{x}^*$ of problem~(\ref{prob.main}),
	, \ie, the optimal distance $\norm{\bm{x}_k-\bm{x}^*}$ is sufficiently small. 
	As seen in Appendix~\ref{appendix.hspg}, 
	it further indicates that the group-sparse optimal solution $\bm{x}^*$ inhabits $\S_k$, and $\S_k$ has already covered the group-support of $\bm{x}^*$, \ie, $\I^{\neq 0}(\bm{x}^*)\subseteq \I^{\neq 0}(\bm{x}_k)$.
	Our goal now becomes minimizing $\psi(\bm{x})$ over $\S_k$ to identify the remaining zero groups, \ie, $\I^0(\bm{x}^*)/\I^{0}(\bm{x}_k)$, which is formulated as the following problem: 
	\begin{equation}\label{prob.half_space_sub_problem}
		\small
		\minimize{\bm{x}\in\S_k}\ \psi(\bm{x})=f(\bm{x})+\lambda r(\bm{x}).
	\end{equation}
	The next iterate $\bm{x}_{k+1}$ is computed as an solution estimate of~problem \eqref{prob.half_space_sub_problem}.
	Particularly, in Algorithm~\ref{alg:main.hspg.outline}, $[\bm{x}_{k+1}]_{\mathcal{I}^0(\bm{x}_k)}\equiv \bm{0}$ will not be updated, and only the entries in $\I^{\neq 0}(\bm{x}_k)$ are free to move. Hence $\psi(\bm{x})$ is smooth on $\S_k$, and~(\ref{prob.half_space_sub_problem}) is a reduced space optimization problem.
	A standard way to solve problem~(\ref{prob.half_space_sub_problem}) would be the stochastic gradient descent equipped with Euclidean projection~\cite{nocedal2006numerical}. However, such a projected method rarely produces zero (group) variables, as the dense Euclidean projected point $\hat{\bm{x}}_{E}\neq \bm{0}$ illustrated in Figure~\ref{figure:half_space_projection}. To address, we introduce a novel half-space projection operator to effectively project an entire group of variables to zeros.
	
	As line~\ref{line:compute_subgrad_psi} and~\ref{line:half_space_start}-\ref{line:half_space_end} in Algorithm~\ref{alg:main.hspg.outline}, we first approximate the (sub)gradient of $\psi$ on the free variables by $[\nu (\bm{x}_k)]_{\I^{\neq 0}(\bm{x}_k)}$, then employ gradient descent over $\I^{\neq 0}(\bm{x}_k)$ to compute a trial point $\widetilde{\bm{x}}_{k+1}$ which is passed into a fresh half-space projection operator $\proj_{\S_k}^{HS}(\cdot)$ defined as 
	\begin{equation}\label{def:proj}
		\left[\proj^{HS}_{\S_k}(\bm{z})\right]_g\coloneqq\bigg\{
		\begin{array}{ll}
			0 & \text{if}\ [\bm{z}]_g^\top[\bm{x}_k]_g< \epsilon\norm{[\bm{x}_k]_g}^2,\\ 
			{[\bm{z}]_g} & \text{otherwise}.
		\end{array}
	\end{equation}
	The above projector of form~(\ref{def:proj}) is not the standard one in Euclidean sense\footnote{ Note that when $r(\bm{x})=\norm{\bm{x}}_1$ where each $g\in\G$ is singleton, then $\S_k$ becomes an orthant face~\cite{chen2020orthant}.}, and it has two advantages: \textit{(i)} the actual search direction $\bm{d}_k:=(\proj^{HS}_{\S_k}(\tilde{\bm{x}}_{k+1})-\bm{x}_k)/\alpha_k$ performs as a descent direction to $\psi(\bm{x}_k)$, \ie, $[\bm{d}_k]_g^\top[\nu(\bm{x}_k))]_g<0$ as $\theta<90^{\circ}$ in Figure~\ref{figure:half_space_projection}, hence the progress to the optimum is made via the sufficient decrease property drawn as Lemma~\ref{lemma:sufficient_decrease_half_space} in Appendix~\ref{appendix.hspg}; then \textit{(ii)} it effectively projects entire groups of variables to zero if the inner product of corresponding entries is sufficiently small. In contrast, the Euclidean projection operator is far away effective to promote group sparsity.
	
	\paragraph{Superiority of~\hspg{} on Group Sparsity Identification.} We now intuitively illustrate the strength of~\hspg{} on group sparsity exploration. In fact, the half-space projection~(\ref{def:proj}) is a more effective sparsity promotion mechanism compared to the standard proximal methods. 
	Particularly, it benefits from a much larger projection region to map a reference point $\hat{\bm{x}}_{k+1}:=\bm{x}_k-\alpha_k\Grad \tilde{f}(\bm{x}_k)$ or its variants to zero. As the 2D case described in Figure~\ref{figure:project_region}, the projection regions of~the state-of-the-art Prox-SG~\cite{duchi2009efficient}, Prox-SVRG~\cite{xiao2014proximal},~\proxspider{}~\cite{zhang2019multi} and SAGA~\cite{defazio2014saga} for~\eqref{prob.main} are $\ell_2$-balls with radius as $\alpha_k\lambda$. In deep learning applications, the step size $\alpha_k$ is usually selected around $10^{-3}$ to $10^{-4}$ or even smaller for convergence. Together with the common setting of $\lambda \ll 1$, their projection regions would vanish rapidly, resulting in the difficulties to produce group sparsity. As a sharp contrast, even though $\alpha_k\lambda$ is near zero, the projection region of~\hspg{} $\{\bm{x}: \bm{x}_k^\top\bm{x}< (\alpha_k\lambda + \epsilon\norm{\bm{x}_k})\norm{\bm{x}_k}\}$ (seen in Appendix~\ref{appendix.hspg}) is still an open half-space which contains those $\ell_2$ balls as well as~\rda~\cite{xiao2010dual}'s if $\epsilon$ is large enough.  
	Conversely, vanilla ADMM alone lacks the mechanism to project a group of variables to zero, unless equips with extra post-processing step~\cite{zhang2018systematic,lin2019toward}.
	In Appendix~\ref{appendix.hspg}, we further reveal that~\hspg{} 
	still maintains the convergence to the optimality as drawn in Theorem~\ref{thm:convergence}. Moreover, we numerically demonstrate the superiority of~\hspg{} in the sense of optimization in Appendix~\ref{appendix.hspg.extensive_experiment}.
	
	\subsection{Pruning Without Fine-Tuning} \label{sec.prune}
	
	The group-sparse solution $\bm{x}^*_{\text{HSPG}}$ over ZIGs to the full model $\mathcal{M}$ is leveraged to construct the slimmer model $\mathcal{M}^*$. 
	Particularly, we prune the redundant structures identified as zero groups $\I^0$ and retain non-zero groups $\I^{\neq 0}$ in $\bm{x}^*_{\text{HSPG}}$.
	Because the parameters of full model are partitioned into ZIGs, the pruned structures contribute none to the model output.
	Therefore, given the same input, the slimmer model $\mathcal{M}^*$ computes the identical output as the full model $\mathcal{M}$ parameterized with $\bm{x}^*_{\text{HSPG}}$.

	\section{Experiment}\label{sec.experiment} 
	
	In this section, we numerically demonstrate the effectiveness of~\algacro{} by one-shot training and pruning without fine-tuning on several benchmark compression tasks for CNNs,~\ie, \vgg{}~\cite{simonyan2014very} for \cifar{}~\cite{Krizhevsky09} and \resnetfifty{}~\cite{he2016deep} for \cifar{}~\cite{Krizhevsky09} and ImagetNet~(\ilsvrc{})~\cite{deng2009imagenet}. We also verify the scalibility of~\algacro{} onto~\bert{}~\cite{NIPS2017_3f5ee243} evaluated on~\squad{}~\cite{rajpurkar2016squad}. All datasets are free to academic usage and do not contain personally identifiable information or offensive content. \cifar{} is under the MIT license, consisting of 50,000 training and 10,000 test images from 10 classes. ImagetNet is a large-scale dataset without license and contains about 1.2 million and 50,000 images in training and validation sets from 1,000 classes. \squad{} is under the CC BY-SA 4.0 license with about 100,000 question/answer pairs splitted into train/dev/test sets as (80/10/10\%). We conduct all experiments on a Nvidia RTX8000 GPU and provide implementation 
	details in Appendix~\ref{appendix.implementation}. 
	
	
	
	
	
	
	\subsection{Deep Convolutional Neural Network}\label{sec.exp.resnet50_imagenet}

	The results on CNN experiments are summarized in Table~\ref{table:vgg_cifar},~\ref{table.resnet50_cifar10} and~\ref{table.resnet50_imagenet}. 
	In particular, we compare \algacro{} to its state-of-the-art counterparts by Top-1/5 accuracy, remaining FLOPs and parameters against the corresponding baseline (full model).
	We report the numbers of other methods based on the corresponding literature and leave as `-' if not reported.
	The best pruning results are marked as bold.
	

	\begin{table}[t]
		\centering
		\caption{\vgg{} and \vggbn{} for \cifar{}. Convolutional layers are in bold.}
		\label{table:vgg_cifar}
		\resizebox{\textwidth}{!}{
			\begin{tabular}{c|c|c|c|c|c}
				\Xhline{3\arrayrulewidth}
				Method & BN & Architecture & FLOPs & \# of Params &  Top-1 Acc. \\
				\hline
				Baseline & \xmark & \textbf{64-64-128-128-256-256-256-512-512-512-512-512-512}-512-512 & 100\% & 100\% & 91.6\% \\
				SBP~\cite{neklyudov2017structured} &  \xmark & \textbf{47-50-91-115-227-160-50-72-51-12-34-39-20}-20-272 & 31.1\% & 5.9\% &  \textbf{91.0\%} \\
				BC~\cite{louizos2017bayesian} &  \xmark & \textbf{51-62-125-128-228-129-38-13-9-6-5-6-6}-6-20 & 38.5\% & 5.4\% &  \textbf{91.0\%} \\
				RBC~\cite{zhou2019accelerate} & \xmark & \textbf{43-62-120-120-182-113-40-12-20-11-6-9-10}-10-22 & 32.3\% & 3.9\% & 90.5\% \\
				RBP~\cite{zhou2019accelerate} & \xmark & \textbf{50-63-123-108-104-57-23-14-9-8-6-7-11}-11-12 & 28.6\% & 2.6\% &  \textbf{91.0\%}\\
				\textbf{\algacro{}} & \xmark & \textbf{21-45-82-110-109-68-37-13-9-7-3-5-8}-170-344  & \textbf{16.3\%} & \textbf{2.5\%} & \textbf{91.0\%}  \\
				\hdashline
				Baseline & \boldcheckmark & \textbf{64-64-128-128-256-256-256-512-512-512-512-512-512}-512-512 & 100\% & 100\% & 93.2\% \\
				EC~\cite{li2016pruning} &  \boldcheckmark &
				\textbf{32-64-128-128-256-256-256-256-256-256-256-256-256}-512-512 &  65.8\% & 37.0\% & 93.1\% \\
				Hinge~\cite{li2020group} & \boldcheckmark & -- & 60.9\% & 20.0\% & 93.6\% \\ 
				SCP~\cite{kang2020operation} & \boldcheckmark & -- & 33.8\% & 7.0\% & \textbf{93.8\%}\\
				\textbf{\algacro{}} & \boldcheckmark & \textbf{22-56-93-123-182-125-95-45-27-21-10-13-19}-244-392 & \textbf{26.8\%}  & \textbf{5.5\%}  & 93.3\% \\
				\hline
				\Xhline{3\arrayrulewidth} 
		\end{tabular}}
		\vspace{-0.5mm}
	\end{table}

	\textbf{\vgg{} for \cifar{}.} 
	We consider the standard~\vgg{} and the version with batch normalization layer after each convolutional layer, referred to as~\vggbn{}. 
	~\algacro{} partitions the parameters into ZIGs following Section \ref{sec.zig}, then trains and prunes the model via~\hspg{}, and finally constructs the slimmer model without fine-tuning. 
	For \vgg{}, as shown in Table~\ref{table:vgg_cifar}, the pruned architecture of \algacro{} indicates that \algacro{} identifies similar redundancy of the intermediate and late convolutional layers compared to other methods, but significantly more of the early convolutional layers. 
	As a result, \algacro{} achieves $83.7\%$ ($1-16.3\%$) FLOPs reduction and $97.5\%$ $(1-2.5\%)$ parameter reduction with the best Top-1 accuracy, which outperforms other state-of-the-arts significantly. 
	For~\vggbn{}, among all, ~\algacro{} reduces FLOPs and parameters to the lowest $26.8\%$ and $5.5\%$, respectively.
	EC~\cite{li2016pruning} and Hinge~\cite{li2020group} 
	achieve the same level of Top-1 accuracy as~\algacro{}, but are substantially outperformed when it comes to FLOPs and parameter reduction. We further present the FLOPs reductions per layer of~\algacro{} in Table~\ref{table.flops_reduction_breakdown} of Appendix~\ref{appendx.flops_breakdown}.
	
	\begin{wraptable}{r}{0.48\textwidth}
		\vspace{-5mm}
		\begin{minipage}{\linewidth}
			\centering
			\caption{\resnetfifty{} for~\cifar{}.}
			\label{table.resnet50_cifar10}
			\resizebox{\linewidth}{!}{
				\begin{tabular}{ c|c|c|c}
					\Xhline{3\arrayrulewidth}
					Method  & FLOPs & \# of Params & Top-1 Acc.\\
					\hline
					Baseline & 100\% & 100\% & 93.5\% \\
					AMC~\cite{he2018amc} & -- & 60.0\% & 93.6\% \\
					ANNC~\cite{yang2020automatic}  & -- & 50.0\% & \textbf{95.0\%} \\
					PruneTrain~\cite{lym2019prunetrain} &  30.0\% & -- & 93.1\% \\ 	   
					N2NSkip~\cite{subramaniam2020n2nskip} & -- & 10.0\% & 94.4\% \\ 
					\textbf{\algacro{}} & \textbf{12.8\%} & \textbf{8.8\%} & 94.4\% \\
					\Xhline{3\arrayrulewidth}
			\end{tabular}}
			\vspace{-4mm}
		\end{minipage}
	\end{wraptable}

	\textbf{\resnetfifty{} for \cifar{}.}  Since~\algacro{} is able to automatically learn a slimmer model of high performance, we compare it with two state-of-the-art automatic neural network compression frameworks, \ie, AMC~\cite{he2018amc} and ANNC~\cite{yang2020automatic}. AMC trains a reinforcement learning agent to predict compression action for each layer environment. ANNC jointly proceeds pruning and quantization within energy constraint. We conduct \algacro{} on their shared experiment,~\ie,~\resnetfifty{} on~\cifar{}. ResNet50 includes both the standard convolutional layers and the layers with residual connections, which are partitioned into ZIGs following Section~\ref{sec.zig}. We report the results in Table~\ref{table.resnet50_cifar10} along with other competitors from~\cite{lym2019prunetrain,subramaniam2020n2nskip}. Based on the results, all methods achieve competitive validation accuracies, where most of them are even higher than the baseline reported in~\cite{he2018amc}. \algacro{} outperforms AMC, ANNC without quantization, PruneTrain and N2NSkip by using only $12.8\%$ FLOPs and 8.8\%  parameters. Note that no FLOPs reduction is reported in~\cite{he2018amc} and~\cite{yang2020automatic}. 
	Finally, we highlight that~\algacro{} is flexible to incorporate quantization as the two techniques are complementary and will leave to future work.

	\begin{wraptable}{r}{0.6\textwidth}
		\vspace{-5mm}
		\begin{minipage}{\linewidth}
			\centering
			\caption{\algacro{} Under Different Switchings ($N=T, 2T, 3T$) for~\vgg{}, \vggbn{} and \resnetfifty{} on \cifar{}}
			\label{table.ablation_study_switching}
			\resizebox{\linewidth}{!}{
				\begin{tabular}{c|c|c|c}
					\Xhline{3\arrayrulewidth}
					Backend  & FLOPs & \# of Params & Top-1 Acc.\\
					\hline
					\vgg{} & 17.0\% $\pm$ 1.4\% & 2.6\% $\pm$ 0.4\% & 90.9\% $\pm$ 0.3\% \\
					\vggbn{} & 25.4\% $\pm$ 1.1\% & 5.0\% $\pm$ 0.5\% & 93.3\% $\pm$ 0.2\% \\
					\resnetfifty{}  & 12.9\% $\pm$ 1.5\%  & 8.5\% $\pm$ 1.0\% & 94.2\% $\pm$ 0.2\% \\
					\Xhline{3\arrayrulewidth}
			\end{tabular}}
			\vspace{-4mm}
		\end{minipage}
	\end{wraptable}
	
	\textbf{Ablation Study on Switching Parameter $\bm N$.} We provide ablation study regarding the impact the switch (parameterized as $N$) between the initialization stage and the group-sparsity stage in Algorithm~\ref{alg:main.outline}. In theory, as shown in Theorem~\ref{thm:convergence} of Appendix~\ref{appendix.hspg.convergence_analysis}, the projection stage should start when the iterate falls nearby a group sparse local minimizer. In practice, we relax it to start the 
	group sparsity stage once the iterate falling into some stationary status regarding the validation accuracy. As described in Appendix~\ref{appendix.implementation.trainindetails}, throughout all experiments, we periodically decay the learning rate per fixed number of epochs parameterized as $T$. At the end of each $T$ epochs, we then proceed a statistical test similar to~\cite{zhang2020statistical} but on the validation accuracy and find that the validation accuracy falls into stationarity near the late epochs of each period. 
	Therefore, in our pruning experiments, we switch to the group-sparsity stage right after the first $T$ epochs. Table~\ref{table.ablation_study_switching} describes the performance of~\algacro{} under varying switching parameters,  from which we observe that~\algacro{} is not largely sensitive to the switching parameter if the group-sparsity stage starts after some stationary condition has been numerically satisfied.

	\begin{wraptable}{r}{0.65\textwidth}
		\vspace{-4mm}
		\begin{minipage}{\linewidth}
			\centering
			\caption{\resnetfifty{} for \imagenet{}.}
			\label{table.resnet50_imagenet}
			\resizebox{\linewidth}{!}{
				\begin{tabular}{ c|c|c|c|c}
					\Xhline{3\arrayrulewidth}
					Method & FLOPs & \# of Params & Top-1 Acc. & Top-5 Acc. \\
					\hline
					Baseline & 100\% & 100\% & 76.1\% & 92.9\% \\
					DDS-26~\cite{huang2018data} & 57.0\% & 61.2\% & 71.8\% & 91.9\% \\
					CP~\cite{He_2017_ICCV} & 66.7\% & -- & 72.3\% & 90.8\% \\		   
					ThiNet-50~\cite{ThiNet_ICCV17} & 44.2\% & 48.3\% & 71.0\% & 90.0\% \\
					RBP~\cite{zhou2019accelerate} & 43.5\% & 48.0\% & 71.1\% & 90.0\%\\
					RRBP~\cite{zhou2019accelerate} & 45.4\% & -- & 73.0\% & 91.0\% \\
					SFP~\cite{he2018soft} & 41.8\% & -- & 74.6\% & 92.1\% \\ 
					Hinge~\cite{li2020group} & 46.6\% & -- & 74.7\% & --\\ 
					GBN-50~\cite{you2019gate} & 44.9\% & 46.6\% & 75.2\% & 92.4\% \\
					GBN-60~\cite{you2019gate} & 59.5\% & 68.2\% & 76.2\% & 92.8\% \\
					Group-HS (2e-5)~\cite{yang2019deephoyer} & 32.4\% & - & 75.2\% & 92.5\% \\
					Group-HS (1e-5)~\cite{yang2019deephoyer} & 52.9\% & - & \textbf{76.4\%} & \textbf{93.1\%} \\
					ResRep~\cite{ding2020lossless} & 45.5\% & - & 76.2\% & 92.9\% \\
					SCP~\cite{kang2020operation} &  45.7\%	& -	& 74.2\% & 92.0\% \\
					\textbf{\algacro{}} & \textbf{34.5\%} & \textbf{35.5\%} & 74.7\% & 92.1\% \\
					\ \ \textbf{\algacro{}}$^*$ & \textbf{34.5\%} & \textbf{35.5\%} & 75.1\% & 92.5\% \\
					\Xhline{3\arrayrulewidth}
				\end{tabular}
			}
		\end{minipage}
	\end{wraptable}
	
	\textbf{\resnetfifty{} for ImageNet.} 
	We now evaluate \algacro{} on ResNet50 for ImageNet. As shown in Table~\ref{table.resnet50_imagenet},~\algacro{} prunes $64.5\% (1-35.5\%)$ parameters to achieve $65.5\% (1-34.5\%)$ FLOPs reduction with only $1.4\%$/$0.8\%$ Top-1/5 accuracy regression compared to the baseline. 
	\algacro{} consistently outperforms the majority of counterparts especially on the FLOPs reduction and the parameter reduction. 
	We note that Hinge~\cite{li2020group} prunes CNNs via structured-sparsity optimization by employing standard stochastic proximal gradient method. It requires several trainings including fine-tuning the pruned model, because it partitions the parameters into non-ZIGs and relies on an empirical truncation mechanism to generate zero groups due to the weakness of proximal operator in deep learning applications~\cite{chen2020orthant}. In contrast, OTO only trains and prunes the full model from scratch once and obtains better pruning results. The comparison between \algacro{} and Hinge stand as evidence of the superiority of \algacro{} due to ZIGs and HSPG. Furthermore, if with more training efforts,~\algacro{} reaches higher Top-1/5 accuracy marked as $^*$ in Table~\ref{table.resnet50_imagenet} and becomes more competitive to stronger competitors, such as GBN~\cite{you2019gate}, Group-HS~\cite{yang2019deephoyer} and ResRep~\cite{kang2020operation}.
	
\begin{figure*}[t!]
		\centering
		\begin{subfigure}[t]{0.49\textwidth}
			\centering
			\includegraphics[width=\linewidth]{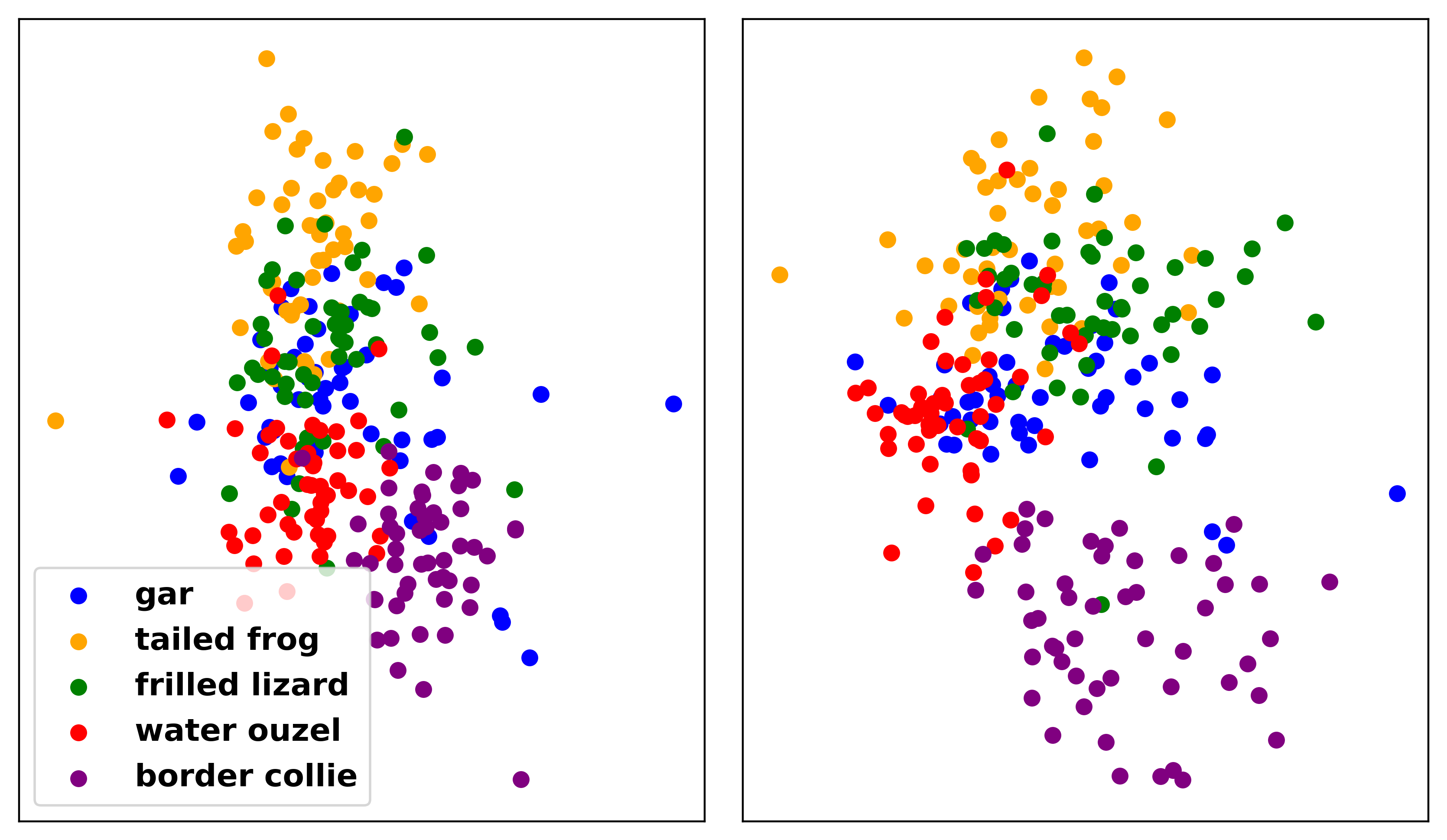}
			\caption{}
		\end{subfigure}%
		~ 
		\begin{subfigure}[t]{0.49\textwidth}
			\centering
			\includegraphics[width=\linewidth]{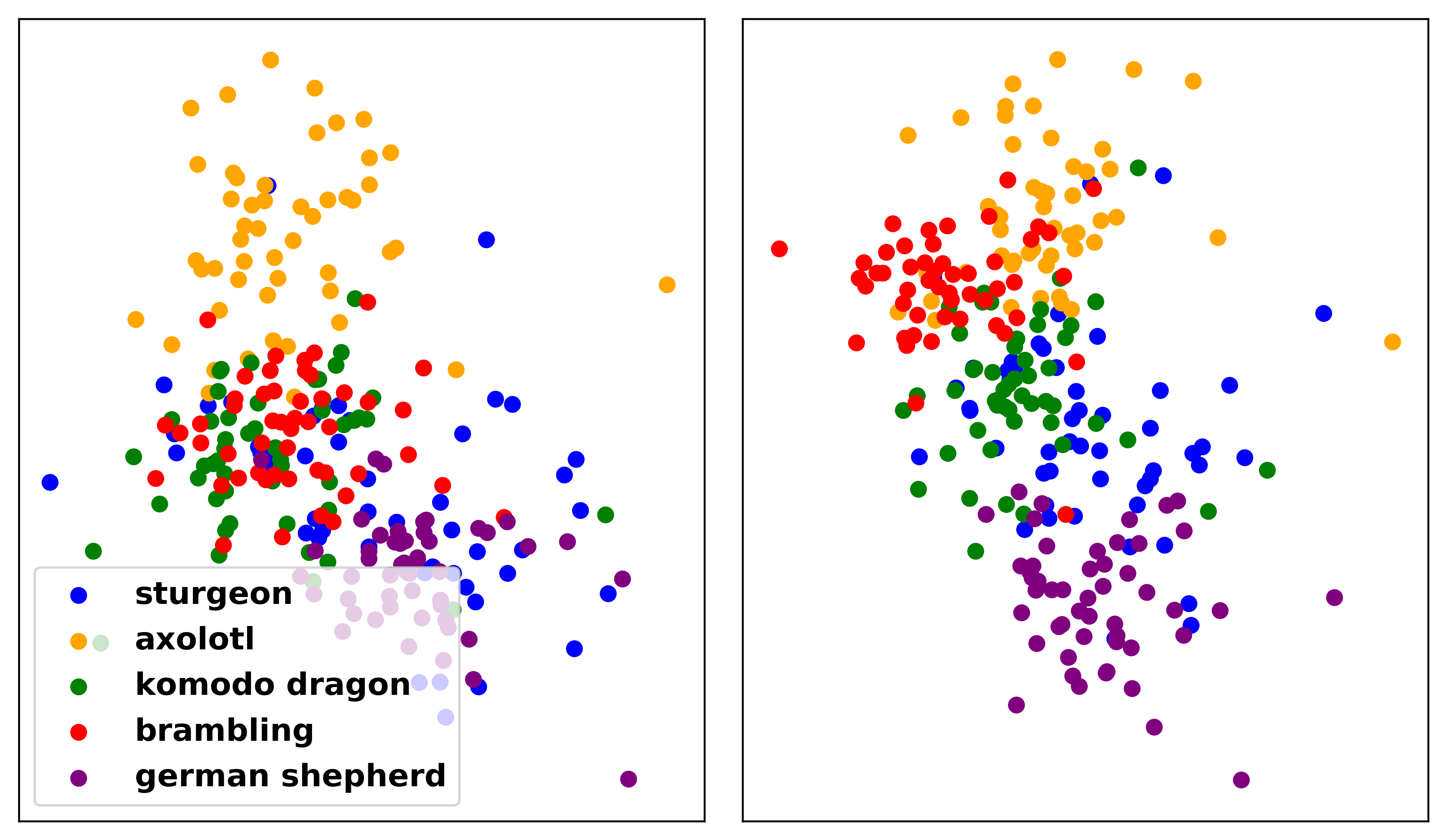}
			\caption{}
		\end{subfigure}
		\caption{Clustering results of ImageNet validation images using deep features extracted by full ResNet50 (left of a and b) and pruned ResetNet50 by \algacro{} (right of a and b). The points are visualized by projecting deep features onto a two-dimensional space via PCA.
		}
		\vspace{-1mm}
		\label{fig:deep-features}
	\end{figure*}

	\textbf{Representation of Deep Features of ImageNet.} It is widely acknowledged that deep neural architectures could be treated as non-linear feature representation extractors. 
	Therefore, 
	we further study the feature representation extracted by~\algacro{} to demonstrate its generalizability to other visual applications
	besides image classification.
	Figure~\ref{fig:deep-features} shows the clustering results of ImageNet validation images using the deep feature extracted by both the baseline ResNet50 and the pruned ResNet50 by \algacro{}.
	Specifically, 
	we extract the deep features over the validation samples in~\imagenet{},~\ie, the tensors fed into the fully connected layer, and project them onto a 2-dimensional space via PCA~\cite{Jolliffe2011}. 
	For illustration, following the hierarchy of~\imagenet{}~\cite{imagenethierychy}, two sets of five classes are randomly selected\footnote{Each selected class belongs to a disjoint upper category.}.
	We observe that the deep features of the pruned ResNet50 by \algacro{} remain structured in the sense that distinct classes are well separated from each other.
	Over all 1000-class ImageNet validation images, \algacro{} achieves 48.2\% clustering accuracy compared to 42.5\% of the baseline ResNet50 using k-means.
	%
	Both observations indicate that with only 35.5\% parameters and 34.5\% FLOPs,
	the pruned ResNet50 is still able to extract highly discriminative deep features. %
	We argue that during model compression, \algacro{} not only achieves parameter and FLOPs reduction, but also preserves the ability of capturing perceptual properties~\cite{zhang2018unreasonable}.
	This is especially important in training and compressing models for 
	many vision tasks, \eg, object detection~\cite{redmon2016you,ren2015faster}, frame interpolation~\cite{bao2019depth,ding2021cdfi,niklaus2017video} and video synthesis~\cite{wang2018video,kwatra2003graphcut}. 
	We leave the application of \algacro{} to broader tasks to future work. 

	\subsection{Large-Scale Transformer}\label{sec.exp.bert_squad}

	We show the scalability of~\algacro{} by pruning the large-scale transformer \bert{}~\cite{NIPS2017_3f5ee243}, evaluated on SQuAD, a question-answering benchmark~\cite{rajpurkar2016squad}. 
	Bert mainly includes embedding layers, fully connected layers and multi-head attention layers. 
	The fully connected layers and the multi-head attention layers are partitioned into ZIGs following Section~\ref{sec.zig}.
	For fair comparisons, we follow the prior Bert compression works~\cite{deleu2021structured,sanh2020movement} and do not prune the embedding layers.
	
	%

	\begin{wraptable}{r}{0.63\textwidth}
		\begin{minipage}{\linewidth}
			\caption{Pruning~Bert on~\squad{}}
			\label{table:bert_squad}
			\centering
			\resizebox{\linewidth}{!}{
				\begin{tabular}{c|c|c|c|c}
					\Xhline{3\arrayrulewidth}
					Method  &\# of Params  & Exact  & F1-score & SpeedUp\\
					\hline
					Baseline &  100\% & 81.0\% & 88.3\% & $1\times$\\
					MaP~\cite{sanh2020movement}  & \textbf{10.0\%} &  67.7\% & 78.5\% & \hspace{1.8mm}$1\times$*\\
					MvP~\cite{sanh2020movement} & \textbf{10.0\%} & 71.9\% & 81.7\% & \hspace{1.8mm}$1\times$*\\
					ProxSSI~\cite{deleu2021structured}  & \hspace{1.3mm}83.4\%$^\dagger$ & 72.3\% & 82.0\% & $1\times$ \\
					\textbf{\algacro{}} & 91.0\% &  \textbf{75.0\%} & \textbf{84.1\%} & $1.1\times$ \\
					\textbf{\algacro{}} & 76.2\% &  72.3\% & 82.1\% & $1.2\times$ \\
					\textbf{\algacro{}} & 66.7\% &  71.9\% & 82.0\% & $1.3\times$ \\
					\textbf{\algacro{}} & 53.3\% &  71.4\% & 81.5\% & $1.5\times$ \\
					\textbf{\algacro{}} & 40.0\% &  70.9\% & 81.1\% & $\bm{1.8\times}$ \\
					\hline%
					\Xhline{3\arrayrulewidth} 
					\multicolumn{5}{l}{* Based on the statement in the official git repository of~\cite{sanh2020movement}.}\\
					\multicolumn{5}{l}{$^\dagger$ Approximate value based on the group sparsity reported in~\cite{deleu2021structured}. }	
				\end{tabular}
			}
		\end{minipage}
	\end{wraptable}
	
	To the best of our knowledge, \algacro{} is the first work that compresses Bert by exploring group sparsity on individual layers and achieves significant parameter reduction and inference speedup\footnote{Knowledge distillation~\cite{hinton2015distilling} and LayerDropout~\cite{fan2019reducing} compresses Bert by pruning layers in their entirety.}.
	In contrast, the existing works~\cite{gordon2020compressing,sanh2020movement,guo2019reweighted} prune individual parameters instead,~\ie, the generated sparsity is not structured. Hence, the computed models typically do not have inference speedup~\cite{sanh2020movement}, unless are executed by specialized hardware and sparse computing library~\cite{han2016eie,chen2018escoin}.
	As shown in Table~\ref{table:bert_squad}, under different group sparsity upper bound constraints,~\algacro{} reduces 9\% to 60\% parameters and achieves up to $1.8\times$ inference speedup based on the average model execution time~\footnote{Run by OnnxRuntime~\cite{onnxruntime}}.  
	In comparison, despite that the pruned model contains $10\%$ parameters, MaP and MvP~\cite{sanh2020movement} do not have any inference speedup.
	On the other hand, the structured sparsity on~\bert{} is studied in~\cite{deleu2021structured} (referred to as ProxSSI), where an adaptive proximal method  is proposed to yield group-sparse solution.
	Nonetheless, ProxSSI optimizes over non-ZIGs and relies on proximal operator to identify group sparsity. Therefore, the groups even parameterized with zeros have to be retained in the model rather than pruned. 
	As a consequence, ProxSSI is not competitive to~\algacro{} on parameter reduction, and there is no reported inference speedup.
	Note that all the pruning methods achieve comparable exact match rate and F1-score.

	\section{Conclusion And Future Work}\label{sec.conclusion}
	
	We propose~\algacro{}, a one-shot deep neural networks (DNNs) training and pruning framework, 
	that compresses full DNNs into slimmer architectures with competitive performances and significant FLOPs and parameter reduction without fine-tuning.
	\algacro{} contains two fundamentals: \textit{(i)} partitions the trainable parameters of DNNs into zero-invariant groups (ZIGs), thereby pruning zero groups does not affect the model output, and \textit{(ii)} trains by a novel optimizer, Half-Space Stochastic Projected Gradient~(\hspg{}), which outperforms proximal methods on group sparsity exploration and maintains comparable convergence. We numerically demonstrate 
	\algacro{} on benchmark experiments,~\ie,~\vgg{} for~\cifar{},~\resnetfifty{} for~\cifar{}/\imagenet{} and~\bert{} for~\squad{}, and achieve state-of-the-art pruning results. We leave automatically generating ZIGs for arbitrary DNNs,
	incorporating quantization and applying \algacro{} to other tasks 
	to future work.

	\newpage
	
	

\appendix

\section{Implementation Details of~\algacro{}}\label{appendix.implementation}

\subsection{ZIG for~\resnetfifty{}}\label{appendix.implementation.zigresnet}

\begin{figure*}[th]
	\centering
	\begin{subfigure}{0.3\linewidth}
		\centering
		\includegraphics[width=0.9\linewidth]{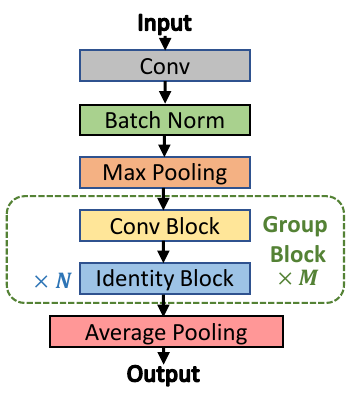}
		\caption{ResNet50.}
		\label{fig:resnet_arch1}
	\end{subfigure}
	\begin{subfigure}{0.65\linewidth}
		\centering
		\includegraphics[width=0.95\linewidth]{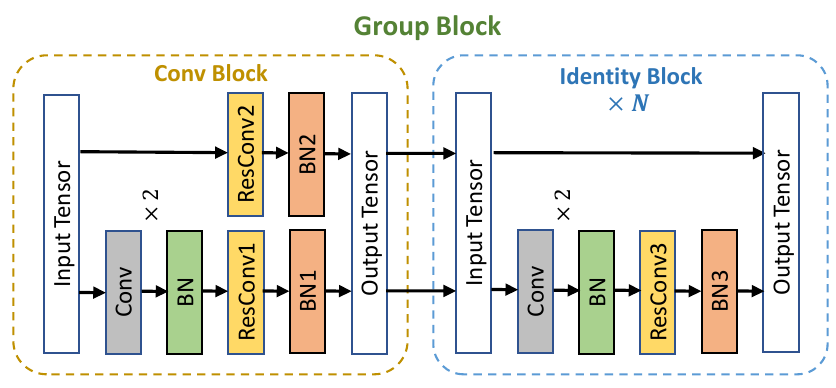}
		\caption{Group block.}\label{fig:resnet_arch2}	
	\end{subfigure}
	\caption{ResNet50 Architecture.}
	\label{fig.resnet50_architecture}
\end{figure*}

Without loss of generality, we illustrate ZIGs for the general ResNet class with \resnetfifty{}. 
As shown in Figure~\ref{fig:resnet_arch1}, ~\resnetfifty{} begins with a Conv-BN, a pooling layer, and extracts features by $M$ Group Blocks that each contains one Conv Block and $N$ Identity Block.
The extracted features are ultimately fed into an average pooling layer for different downstream computations.
There exist two types of convolution structures inside~\resnetfifty{}: \textit{(i)} the regular Conv-BN (see Section~\ref{sec.zig}), marked as gray and green blocks in Figure~\ref{fig:resnet_arch2}, and \textit{(ii)} the ResConv-BN of which the output shares the same dimension with another ResConv-BN, marked as yellow and brown in Figure~\ref{fig:resnet_arch2}.


For~\resnetfifty{}, we partition regular Conv-BN following Section~\ref{sec.zig}.
For ResConv-BN, within each Group Block, the intermediate input/output tensors in Conv/Indentity Blocks share the same dimension, and hence all the ResConv-BNs in one Group block share the same number of 3D filters. 
Consequently, their flattened filter matrices has the same number of rows.
Figure~\ref{fig:resnet_arch2} breaks down the architecture of a Group Block.
The output tensors of ResConv-BN1 and ResConv-BN2 in Conv Block, denoted as $\bm{\O}^{1}$ and $\bm{\O}^{2}$, are computed by~\eqref{eq:resnet_o1} and~\eqref{eq:resnet_o2} respectively.
They are then summed up as the input tensor of the subsequent identify block $\bm{\I}^{I_1}$, indicating that $\bm{\O}^{1}$ and $\bm{\O}^{2}$ have the same shape and their flattened filter matrices $\small \bm{\hat{\K}}^{1}$ and $\small \bm{\hat{\K}}^{2}$ has the same number of rows.
As~\eqref{eq:resnet_i_i2}, $\bm{\I}^{I_1}$ later sums the output tensor of ResConv-BN3 $\bm{\mathcal{O}}^{3}$ to yield the input tensor to the next Identity Block $\bm{\mathcal{I}}^{I_2}$, implying the filter matrix of ResConv-BN3 $\small \bm{\hat{\K}}^{3}$ has the same number of rows as $\small \bm{\hat{\K}}^{1}$ and $\small \bm{\hat{\K}}^{2}$.

\begin{align}
	\bm{\mathcal{O}}^{1}&\gets \frac{a(\bm{\mathcal{I}}^1 \otimes \bm{\hat{\K}}^{1} +\bm{b}^1)-\bm{\mu}^1}{\bm{\sigma}^{1}}\odot\bm{\gamma}^1 +\bm{\beta}^1\label{eq:resnet_o1}\\
	\bm{\mathcal{O}}^{2}&\gets \frac{a(\bm{\mathcal{I}}^2 \otimes \bm{\hat{\K}}^{2} +\bm{b}^2)-\bm{\mu}^2}{\bm{\sigma}^{2}}\odot\bm{\gamma}^2 +\bm{\beta}^2\label{eq:resnet_o2}\\
	\bm{\mathcal{I}}^{I_1}&\gets \bm{\mathcal{O}}^{1} + \bm{\mathcal{O}}^{2}\\
	\bm{\mathcal{O}}^{3}&\gets \frac{a(\bm{\mathcal{I}}^3 \otimes \bm{\hat{\K}}^{3} +\bm{b}^3)-\bm{\mu}^3}{\bm{\sigma}^{3}}\odot\bm{\gamma}^3 +\bm{\beta}^3\\
	\bm{\mathcal{I}}^{I_2}&\gets \bm{\mathcal{I}}^{I_1} + \bm{\mathcal{O}}^{3}\label{eq:resnet_i_i2}
\end{align}

\begin{figure}[h]
	\centering
	\includegraphics[width=0.65\linewidth]{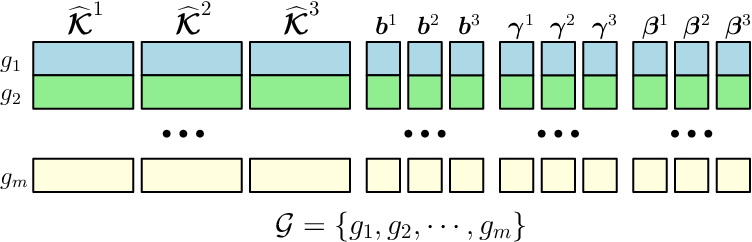}
	\caption{Zero Invariant Groups for the three ResConv-BN of a Group Block.}
	\label{fig.resnet50_zig}
\end{figure}

Therefore, based on~\eqref{eq:resnet_o1} to~\eqref{eq:resnet_i_i2}, to make the entire Group Block zero-invariant, we group each $i^{th}$ row of the filter matrix for all the Res-Conv-BNs of same group block.
In doing so, any one row of parameters being zeros results in the output, \ie, the corresponding channel of feature map, being zeros.
Figure~\ref{fig.resnet50_zig} shows ZIG for the three ResConv-BN of a Group Block.
Regardless of the input, the $i^{th}$ channel-wise matrix of $\bm{\I}^{I_1}$ are zeros if and only if both $i^{th}$ channel-wise matrices of $\bm{\O}^{1}$ and $\bm{\O}^{2}$ are equal to zero.
This is equivalent to both $i^{th}$ rows of $\small \bm{\hat{\K}}^{1}$ and $\small \bm{\hat{\K}}^{2}$ being zeros.
Similarly, $i^{th}$ channel-wise matrix of $\bm{\I}^{I_2}$ being zeros regardless of the input further requires the $i^{th}$ row of $\small \bm{\hat{\K}}^{3}$ to be grouped in the ZIG. 
%


\subsection{Training Details}\label{appendix.implementation.trainindetails}

We implement~\algacro{} in PyTorch. The key ingredient~\hspg{} is packaged as an optimizer class which is flexible to various applications. During the experiment, the trainable parameters of the full model $\M$ are firstly partitioned into a ZIG set $\G$ wherein each group is tagged as its corresponding atomic layer category,~\eg, fully-connected layer or convolutional layer. The ZIG set $\mathcal{G}$ is treated as an argument to the~\hspg{} constructor. Then $\M$ is trained from scratch by~\hspg{} where the group-wise variables are updated based on their tagged layer category. 
In our repository, we provide the prescribed ZIG partitions for the DNNs used in this paper,~\ie,~\vgg{},~\vggbn{},~\resnetfifty{} and~\bert{}. 
For other models, one can easily follow Section~\ref{sec.zig} and Appendix~\ref{appendix.implementation.zigresnet} to construct a ZIG partition and feed it as an argument to the HSPG optimizer.
After training, a full group-sparse model with high performance is achieved. Finally, a slimmer pruned model $\M^*$ is constructed following~Section~\ref{sec.prune} without fine-tuning and has the identical performance as the full group-sparse model. 
We provide the implementation in \url{https://github.com/tianyic/only_train_once} and the pruned models 
associated with the corresponding full group sparse models in \url{https://tinyurl.com/otocheckpoints}. 

\paragraph{Parameter Settings.} We conduct all experiments on an Nvidia RTX8000 graphics card with 48 GB memory. For all CNN experiments, the step size (learning rate) $\alpha_k$ is initialized as $10^{-1}$, and decayed by a factor 0.1 periodically $T$ epochs till the minimum value $10^{-4}$. The selection of $T$ depends on the max number of epochs $K$.
We follow various benchmark online resources to select $K$. Particularly, for all~\cifar{} experiments, we follow the model pre-training settings in~\cite{li2020group} and set $K$ as 300. Note that by using the same number of epochs,~\algacro{} achieves both slimmer model and competitive performance simultaneously. 
For the \imagenet{} experiment, following~\cite{he2016deep}, we set $T$ as 30 and $K$ as 120. For all \bert{} experiments, the step size $\alpha_k$ is initialized as $10^{-1}$ and decayed by a factor 0.1 after 4 epochs to be as $10^{-2}$. 

We set the mini-batch size as the commonly used 64 for~\cifar{}, 256 for~\imagenet{} and 32 for~\squad{} experiments. For all experiments, we initialize the regularization coefficient $\lambda$ as $10^{-3}$ to balance between model performance and group sparsity. In particular, $\lambda$ as $10^{-3}$ is the maximum value from the candidate set $\{10^{-2},10^{-3},10^{-4}\}$ which returns competitive evaluation results to the majority of the tested models trained without regularization. In addition, to favor more on the model performance, if group sparsity becomes stationary, we decay $\lambda$ by a factor 0.1 periodically after stepping into Group-Sparsity Stage. The control parameter $\epsilon\in[0,1)$ in the half-space projection~\eqref{def:proj} controls the aggressiveness level of group sparsity promotion, which is typically fixed as $0$ since for most of experiments, $\epsilon\equiv 0$ has resulted in sufficiently good experiment results. In case if group sparsity is not sufficiently yielded, we provide an adaptive mechanism to  increase $\epsilon$ by 0.1 till the upper bound 0.999. 
For the setting of $N$ which controls when switching to the Group-Sparsity Stage, we proceed a test on objective value stationarity similarly to~\cite[Section 2.1]{zhang2020statistical} and empirically set $N\equiv T$ for CNN experiments since the validation accuracy values become stationary at the late epochs till $T$. Hence, the Group-Sparsity Stage starts after $T$ epochs and is accompanied with the $\alpha_k$ decaying. For Bert experiment, we empirically set $N$ as 1 since the F1-score and exact match rate becomes stationary after one epoch training.  

\paragraph{Additional Remarks.} For the ZIG partition of~\resnetfifty{} on~\cifar{}, we include all trainable variables of~\resnetfifty{} and apply the ZIG partition described in Appendix~\ref{appendix.implementation.zigresnet} for ResConv-BN and the ZIG partition described in Section~\ref{sec.zig} for standard Conv-BN. For the ZIG partition of~\resnetfifty{} on~\imagenet{}, we construct ZIGs for standard Conv-BN only. This is because we observe that ZIG partition for ResConv-BN lead to accuracy regression in spite of more FLOPs reduction, (15\% FLOPs with up to 71\% Top-1 Accuracy). The cause is that it decreases the number of features maps generated by the entire Group Block. Additionally, for Bert experiments, to study the accuracy evolution against different compression rates, we set extra constraints to bound the maximum group sparsity ratio,~\eg, 30\%, 50\%, 70\%, and do not yield new zero groups if the upper bound has been exceeded. Note that without any constraint,~\algacro{} reaches about 95\% group sparsity ratio with 80\% F1-score.

\subsection{Error Bar Analysis}\label{appendix.implementation.error_bar}
In this section, we report the overall statistics of the experiments and analyze the error bar.
We note that for fair comparison with others, in the main body of paper, we report the best results in terms of remaining FLOPs/parameters and Top-1/5 accuracy. 
We conduct all experiments three times with different random seeds.

\begin{table}[h]
	\centering
	\caption{\algacro{} for CNN Experiments  (mean $\pm$ std)}
	\label{table:error_analysis_cnn}
	\centering
	\resizebox{0.75\columnwidth}{!}{
		\begin{tabular}{c|c|c|c|c}
			\Xhline{3\arrayrulewidth}
			Model & Dataset  & FLOPs & \# of Params  & Top-1 Acc.  \\
			\hline
			\vgg{} & \cifar{} &  $16.9\%\pm 1.5\%$ & $2.7\%\pm 0.2\%$ & $90.7\%\pm 0.3\%$ \\
			\vggbn{} & \cifar{}  & $26.9\%\pm 0.1\%$ &  $5.5\%\pm 0.1\%$ & $93.2\%\pm 0.2\%$ \\
			\resnetfifty{} & \cifar{} & 
			$11.9\%\pm 1.7\%$ 
			& 
			$8.8\%\pm 0.4\%$ 
			& $93.9\%\pm 0.5
			\%$ \\
			\resnetfifty{} & \imagenet{} & $34.8\%\pm 1.8\%$ & $35.9\%\pm 1.7\%$ & $73.3\%\pm 1.1\%$ \\
			\hline%
			\Xhline{3\arrayrulewidth} 
		\end{tabular}
	}
\end{table}

Training neural networks is equivalent to solving a non-convex optimization problem which has numerous local minimizers, thereby training from scratch like~\algacro{} may generate solutions close to stationary points with different attributes. As shown in~Table~\ref{table:error_analysis_cnn}, we can see that for the CNN experiments,~\algacro{} performs reliably to achieve significant FLOPs and parameters reduction and competitive Top-1 accuracy with small fluctuations. 

\subsection{FLOPs Reduction Breakdown}\label{appendx.flops_breakdown}

We provide the layer-wise FLOPs reduction for~\vgg{} on~\cifar{}. As shown in Table~\ref{table.flops_reduction_breakdown}, the majority of the FLOPs reduction via OTO comes from a few middle ConvLayers (over 10\% to the overall FLOPs reductions) instead of the first ConvLayer (0.45\% to the overall FLOPs reduction). In general, the distribution of FLOPs reduction per Layer of OTO is similar to other pruning baselines. 

\begin{table}[h]
	\centering
	\caption{FLOPs Reduction Breakdown for the ConvLayers of VGG16 on CIFAR10}
	\label{table.flops_reduction_breakdown}
	\resizebox{0.8\columnwidth}{!}{
		\begin{tabular}{ccccc}
			\Xhline{3\arrayrulewidth}
			
			\multirow{2}{*}{ConvLayer Index} &  \multicolumn{2}{c}{\# of Output Channels}  & \multicolumn{2}{c}{FLOPs Reduction} \\
			& Original & Pruned &  Quantity (Million) & Percentage (\%)\\
			\hline
			1 & 64 & 21 & 1.19M & 0.45\% \\
			2 & 64 & 45& 29.04M & 11.07\% \\
			3 & 128 & 82 & 10.47M & 3.99\% \\
			4 & 128 & 110 & 17.22M & 6.57\% \\
			5 & 256 & 109 & 11.97M & 4.56\% \\
			6 & 256 & 68 & 33.48M & 12.77\% \\
			7 & 256 & 37 & 36.30M & 13.84\% \\
			8 & 512 & 13 & 18.81M & 7.17\% \\
			9 & 512 & 9 & 37.73M & 14.38\% \\
			10 & 512 & 7 & 37.74M & 14.39\% \\
			11 & 512 & 3 & 9.44M & 3.60\% \\
			12 & 512 & 5 & 9.44M & 3.60\% \\
			13 & 512 & 8 & 9.44M & 3.60\% \\
			\Xhline{3\arrayrulewidth}
		\end{tabular}
	}
\end{table}

\section{Convergence Analysis of \hspg{} }\label{appendix.hspg}

In this section, we provide theoretical analysis of HSPG. We focus on the most popular setting of optimization problem \eqref{prob.main} as follows
\begin{equation}\label{prob.main-appendix}
	\minimize{\bm{x}\in\mathbb{R}^n}\ \psi(\bm{x}):=f(\bm{x})+\lambda r(\bm{x}),\ f(\bm{x}) := \frac{1}{N} \sum_{i = 1}^N f_i (\bm{x}), 
\end{equation}
Here $f(\bm{x})$ is defined as the average of $N$ task-specific loss functions $f_i: \mathbb{R}^n \mapsto \mathbb{R}, ~ \forall ~ i = 1, \ldots, N$. The stochastic gradient $\nabla \tilde{f}$ proposed in Section~\ref{sec.hspg} can be obtained via a uniformly chosen mini-batch $\B \subseteq [N]$ as follows: for any $\bm{x} \in \mathbb{R}^n$, given $\B$, we have
\begin{align}
	\nabla \tilde{f} (\bm{x})= \nabla \bigg( \underbrace{ \frac{1}{|\B|} \sum_{i \in \B} f_i(\bm{x})}_{=: f_{\B}(\bm{x})} \bigg),
\end{align}
in short, we denote above term as $\nabla f_{\B}(\bm{x})$ where $f_{\B} (\bm{x})$ is the average of loss functions with repsect to mini-batch $\B$. Similarly, let $\psi_{\B}(\bm{x}) := f_{\B} (\bm{x}) + \lambda r(\bm{x})$ for all $\bm{x} \in \mathbb{R}^n$. 

\textbf{Organization.} The Section~\ref{appendix.hspg} is organized as follows: From Section~\ref{appendix.hspg.related_work} to Section~\ref{appendix.hspg.initial_sage}, we present the convergence result and the sparse recovery guarantee for \halfspacestep{}. More specifically, 
\begin{itemize}
	\item In Section~\ref{appendix.hspg.related_work}, we first presented the existing related work of solving the problem~\eqref{prob.main-appendix}. 
	\item In Section~\ref{appendix.hspg.sufficient_decrease_half_space}, we show the sufficient decrease of \halfspacestep{} under Assumption~\ref{assumption:hspg-sufficient-decrease}. \item In Section~\ref{appendix.hspg.project_region}, we derive the projection region of \halfspacestep{} and compare this projection region with existing methods. 
	\item In Section~\ref{appendix.hspg.convergence_analysis}, we give the convergence result of \halfspacestep{} as stated in Theorem~\ref{thm:convergence} under the Assumption~(\ref{assumption:hs-obj-property}, \ref{assumption:stepsize}). 
\end{itemize}
To complete the story, in Section~\ref{appendix.hspg.initial_sage}, we show that the ``close enough'' condition required in Theorem~\ref{thm:convergence} can be achieved by the \textit{Sub-gradient Descent Step} under the Assumption~\ref{assumption:init-convergence}. Moreover, we further point out that: (1) the \textit{Sub-gradient Descent Step} we used to achieve a ``close enough'' solution can be replaced by other methods, and (2) the Assumption~\ref{assumption:opt-solution-property} is only a sufficient condition that we could use to show the ``close enough'' condition. 

\subsection{Related Work}\label{appendix.hspg.related_work}

Problem~(\ref{prob.main-appendix}) has been well studied in deterministic optimization with various algorithms that are capable of returning solutions with both low objective value and high group sparsity under proper $\lambda$~\cite{yuan2006model,roth2008group,huang2011learning,ndiaye2017gap}. Proximal methods are classical approaches to solve the structured non-smooth optimization~\eqref{prob.main-appendix}, including the popular proximal gradient method (Prox-FG) which only uses the first-order derivative information. When $N$ is huge, stochastic methods become ubiquitous to operate on a small subset to avoid the costly evaluation over all instances in deterministic methods for large-scale problems. Proximal stochastic gradient method~(\proxsg)~\cite{duchi2009efficient} is the natural stochastic extension of Prox-FG. Regularized dual-averaging method (\rda)~\cite{xiao2010dual,yang2010online} is proposed by extending the dual averaging scheme in~\cite{nesterov2009primal}. To improve the convergence rate, 
there exists a set of incremental gradient methods inspired by  SAG~\cite{roux2012stochastic} to utilizes the average of accumulated past gradients. For example, 
proximal stochastic variance-reduced gradient method (\proxsvrg{})~\cite{xiao2014proximal} and proximal spider (\proxspider)~\cite{zhang2019multi} are developed to adopt multi-stage
schemes based on the well-known variance reduction technique SVRG proposed in~\cite{johnson2013accelerating} and Spider developed in~\cite{fang2018spider} respectively.  \saga{}~\cite{defazio2014saga} stands as the midpoint between SAG and Prox-SVRG.

Compared to deterministic methods, the studies of structured sparsity regularization~(\ref{prob.main-appendix}) in stochastic field become somewhat rare and limited. \proxsg{},~\rda{},~\proxsvrg{}, Prox-Spider and~\saga{} are valuable state-of-the-art stochastic algorithms for solving problem~(\ref{prob.main-appendix}) but with apparent weakness. Particularly, these existing stochastic algorithms typically meet difficulties to achieve both decent convergence and effective group sparsity identification simultaneously (\eg, small function values but merely dense solutions), because of the randomness and the limited sparsity-promotion mechanisms. In depth,~\proxsg{},~\rda{},~\proxsvrg{},~\proxspider{} and~\saga{} derive from proximal gradient method to utilize the proximal operator to produce group of zero variables. Such operator is generic to extensive non-smooth problems, consequently perhaps not sufficiently insightful if the target problems possess certain properties,~\eg, the group sparsity structure as problem~(\ref{prob.main-appendix}). In fact, in convex setting, the proximal operator suffers from variance of gradient estimate; and in non-convex setting, especially deep learning, the discreet step size (learning rate) further deteriorates its effectiveness on the group sparsity promotion, as shown in Section~\ref{sec.hspg} of the main body that the projection region vanishes rapidly except~\rda{}. \rda{} has superiority on finding manifold structure to others~\cite{lee2012manifold}, but inferiority on the objective convergence.
Besides, the variance reduction techniques are typically required to measure over a huge mini-batch data points in both theory and practice which is probably prohibitive for large-scale problems, and have been observed as sometimes noneffective for deep learning applications~\cite{defazio2019ineffectiveness}. On the other hand, to introduce sparsity, there exist heuristic weight pruning methods~\cite{li2016pruning,luo2017thinet}, whereas they commonly do not equip with theoretical guarantee, so that easily diverge and hurt generalization accuracy.

\subsection{Sufficient Decrease of~\halfspacestep{}}\label{appendix.hspg.sufficient_decrease_half_space}

Before we present the convergence result of \halfspacestep{} to the global group-sparsity solution, in this part, we first show that the sufficient decrease property holds for \halfspacestep{} under the following Assumption~\ref{assumption:hspg-sufficient-decrease}. 
\begin{assumption} \label{assumption:hspg-sufficient-decrease}
	Assume the following assumptions hold. 
	\begin{itemize}
		\item \textbf{(A\ref{assumption:hspg-sufficient-decrease}-1).} $f: \mathbb{R}^n \mapsto \mathbb{R}$ is differentiable and $L$ smooth.
		
		\item \textbf{(A\ref{assumption:hspg-sufficient-decrease}-2).} $r: \mathbb{R}^n \mapsto \mathbb{R}$ is sub-differentiable and convex.
		
		\item \textbf{(A\ref{assumption:hspg-sufficient-decrease}-3).} $\psi = f + \lambda r: \mathbb{R}^n \mapsto \mathbb{R}$ is sub-differentiable over all points $\bm{x} \in \mathbb{R}^n$.
	\end{itemize}
\end{assumption}
For any $k > N_{\mathcal{P}}$ (in \halfspacestep{} of Algorithm~\ref{alg:main.hspg.outline}), recall the next iterate $\bm{x}_{k + 1}$ and the search direction
\begin{align}
	\bm{d}_k := \frac{\bm{x}_{k + 1} - \bm{x}_{k}}{\alpha_k} = \frac{\text{Proj}_{\S_k}^{\text{HS}}(\bm{x}_k-\alpha_k\Grad \psi_{\B_k}(\bm{x}_k)) - \bm{x}_{k}}{\alpha_k}.
\end{align}
Define
\begin{align}
	\hat{\mathcal{G}}_k:= & ~ \I^{\neq 0}(\bm{x}_k)\cap \I^{0}(\bm{x}_{k+1}) \label{def:hat_G}\\
	\tilde{\mathcal{G}}_k:= & ~ \I^{\neq 0}(\bm{x}_k)\cap \I^{\neq 0}(\bm{x}_{k+1})\label{def:tilde_G}
\end{align}
be the sets of groups which projects or not onto zero. We claim that the following Lemma~\ref{lemma:sufficient_decrease_half_space} holds. 
\begin{lemma}\label{lemma:sufficient_decrease_half_space} Under Assumption~\ref{assumption:hspg-sufficient-decrease}, the search direction $\bm{d}_k$ is a descent direction for $\psi_{\B_k}(\bm{x}_k)$, \ie, $\bm{d}_k^\top\Grad \psi_{\B_k}(\bm{x}_k)<0$. Moreover, we have the following sufficient decrease property holds, 
	\begin{equation}
		\small
		\begin{split}
			\psi_{\mathcal{B}_k}(\bm{x}_{k+1})\leq & \psi_{\mathcal{B}_k}(\bm{x}_{k})-\left(\alpha_k-\frac{\alpha_k^2L}{2}\right)\sum_{g\in\tilde{\G}_k}\norm{[\Grad \psi_{\mathcal{B}_k}(\bm{x}_{k})]_g}^2-\left(\frac{1-\epsilon}{\alpha_k}-\frac{L}{2}\right)\sum_{g\in\hat{\G}_k}\norm{[\bm{x}_{k}]_g}^2.
		\end{split}
	\end{equation}
	\vspace{-0.4cm}
\end{lemma}
\begin{proof}
	\textbf{Proof of Descent Direction.} 
	It follows the~\halfspacestep{} in Algorithm~\ref{alg:main.hspg.outline} and the definition of $\tilde{\G}_k$ and $\hat{\G}_k$ as~(\ref{def:tilde_G}) and~(\ref{def:hat_G}) that $\bm{x}_{k+1}=\bm{x}_k+\alpha_k\bm{d}_k$ where $\bm{d}_k$ is
	\begin{equation}\label{eq:proof_d_k_def}
		[\bm{d}_k]_g=
		\begin{cases}
			-[\nabla \psi_{\mathcal{B}_k}(\bm{x}_{k})]_g& \text{if}\ g\in\tilde{\mathcal{G}}_k=\I^{\neq 0}(\bm{x}_k)\bigcap \I^{\neq 0}(\bm{x}_{k+1}),\\
			-[\bm{x}_{k}]_g/\alpha_k & \text{if}\ g \in \hat{\mathcal{G}}_k=\I^{\neq 0}(\bm{x}_k)\bigcap \I^{0}(\bm{x}_{k+1}),\\
			0 & \text{otherwise}.
		\end{cases}
	\end{equation}
	We also notice that for any $g\in\hat{\mathcal{G}}_k$, the following holds
	\begin{equation}\label{eq:descent_direction_tmp1}
		\begin{split}
			[\bm{x}_k-\alpha_k\nabla \psi_{\mathcal{B}_k}(\bm{x}_k)]_g^\top[\bm{x}_k]_g<\epsilon \norm{[\bm{x}_k]_g}^2,\\
			(1-\epsilon)\norm{[\bm{x}_k]_g}^2< \alpha_k[\nabla \psi_{\mathcal{B}_k}(\bm{x}_k)]_g^\top[\bm{x}_k]_g.
		\end{split}
	\end{equation}
	For simplicity, let $\I^{\neq 0}_k:=\I^{\neq 0}(\bm{x}_k)$. Since $[\bm{d}_k]_g=\bm{0}$ for any $g\in \I^{0}(\bm{x}_k)$, then by~(\ref{eq:proof_d_k_def}) and~(\ref{eq:descent_direction_tmp1}), we have 
	\begin{equation} \label{eq:d_k-grad_psi-product}
		\begin{split}
			\bm{d}_k^\top\nabla \psi_{\B_k}(\bm{x}_k)&=[\bm{d}_k]_{\I^{\neq 0}_k}^\top[\nabla \psi_{\B_k}(\bm{x}_k)]_{\I^{\neq 0}_k}\\
			&=-\sum_{g\in\tilde{\G}_k}\norm{[\nabla \psi_{\B_k}(\bm{x}_k)]_g}^2-\sum_{g\in \hat{\G}_k}\frac{1}{\alpha_k}[\bm{x}_k]_g^\top[\nabla \psi_{\B_k}(\bm{x}_k)]_g\\
			&\leq -\sum_{g\in\tilde{\G}_k}\norm{[\nabla \psi_{\B_k}(\bm{x}_k)]_g}^2-\sum_{g\in \hat{\G}_k}\frac{1}{\alpha_k^2}(1-\epsilon)\norm{[\bm{x}_k]_g}^2< 0,
		\end{split}
	\end{equation}
	holds for any $\epsilon\in[0,1)$, which implies that $\bm{d}_k$ is a descent direction for $\psi_{\B_k}(\bm{x}_k)$. 
	
	\textbf{Proof of Sufficient Decrease.}  Now, we start to prove the suffcient decrease of~\halfspacestep{}. By assumption, $f: \mathbb{R}^n \mapsto \mathbb{R}$ is $L$ smooth and $r: \mathbb{R}^n \mapsto \mathbb{R}$ is convex. Therefore
	\begin{align}
		& ~ \psi_{\B_k}(\bm{x}_{k}+\alpha_k \bm{d}_{k}) \\
		= & ~ f_{\B_k}(\bm{x}_{k}+\alpha_k \bm{d}_{k}) + \lambda r (\bm{x}_{k}+\alpha_k \bm{d}_{k}) \\
		\leq & ~ f_{\B_k}(\bm{x}_{k}) + \alpha_k \bm{d}_{k}^{\top} \nabla f_{\B}(\bm{x}_{k}) + \frac{ \alpha_k^2 L}{2} \|\bm{d}_{k}\|^2 & \text{by Assumption~\ref{assumption:hspg-sufficient-decrease}}\\
		& ~ + \lambda r(\bm{x}_{k}) + \alpha_k \lambda \bm{d}_{k}^{\top} \zeta(\bm{x}_{k}) \\
		= & ~ \psi_{\B_k}(\bm{x}_{k}) + \alpha_k \bm{d}_k^\top\nabla \psi_{\B_k}(\bm{x}_k) + \frac{\alpha_k^2 L}{2} \|\bm{d}_{k}\|^2 \\
		\leq & ~ \psi_{\mathcal{B}_k}(\bm{x}_{k})-\left(\alpha_k-\frac{\alpha_k^2L}{2}\right)\sum_{g\in\tilde{\G}_k}\norm{[\nabla \psi_{\mathcal{B}_k}(\bm{x}_{k})]_g}^2 & \text{by inequality \eqref{eq:d_k-grad_psi-product} \& $\bm{d}_k$ definition} \\
		& ~ -\left(\frac{1-\epsilon}{\alpha_k}-\frac{L}{2}\right)\sum_{g\in\hat{\G}_k}\norm{[\bm{x}_{k}]_g}^2,
	\end{align}
	which completes the proof. 
\end{proof}

According to Lemma~\ref{lemma:sufficient_decrease_half_space}, the objective value $\psi_{\B}(\bm{x})$ with $\mathbb{E}[\psi_{\B}(\bm{x})|\bm{x}] = \psi(\bm{x})$ achieves a sufficient decrease in Half-Space Step given $\alpha_k$ is small enough. Taking the expectation over mini-batch $\B$ on both sides, it is straight-forward to obtain the expectation version of the sufficient decrease property.

\begin{corollary}\label{corollary:psi_epoch_decrease}	
	Similarly, under Assumption~\ref{assumption:hspg-sufficient-decrease}, for all $k > N_{\mathcal{P}}$, we have
	\begin{equation}
		\psi(\bm{x}_{k+1}) \leq \psi(\bm{x}_k)-\sum_{g\in\tilde{\mathcal{G}}_k}\left(\alpha_k-\frac{\alpha_k^2L}{2}\right)\mathbb{E}\left[\norm{[\nabla \psi_{\mathcal{B}_k}(\bm{x}_k) ]_g }^2\right]-\left(\frac{1-\epsilon}{\alpha_k}-\frac{L}{2}\right)\sum_{g\in\hat{\G}_k}\norm{[\bm{x}_{k}]_g}^2.
	\end{equation}
\end{corollary}

\subsection{Projection Region of \halfspacestep{}}\label{appendix.hspg.project_region}

In this part, we derive the projection region of~\halfspacestep{}, and reveal that is a superset of the projection region of existing methods, e.g.  ~\proxsg{},~\proxsvrg{} and~\proxspider{}, under the same $\alpha_k$ and $\lambda$.
\begin{proposition}
	For any $k > N_{\mathcal{P}}$, given $\bm{x}_k$, the next iterate $\bm{x}_{k + 1}$ obtained by the \halfspacestep{} satisfies that: for any group $g \in\mathcal{I}^{\neq 0}(\bm{x}_k)$, 
	\begin{align}
		[\bm{x}_{k+1}]_g=
		\begin{cases}
			[\hat{\bm{x}}_{k+1}]_g-\alpha_k\lambda \frac{[\bm{x}_k]_g}{\norm{[\bm{x}_k]_g}}& \text{if }\ [\hat{\bm{x}}_{k+1}]_g^\top[\bm{x}_k]_g> (\alpha_k\lambda + \epsilon)\norm{[\bm{x}_k]_g},\\
			0 & \text{otherwise},
		\end{cases}
	\end{align}
	where $\hat{\bm{x}}_{k+1} := \bm{x}_k-\alpha_k\Grad f_{\mathcal{B}_k}(\bm{x}_k)$. Moreover, we claim that if $\norm{[\hat{\bm{x}}_{k+1}]_g}\leq \alpha_k\lambda$,  then $[\bm{x}_{k+1}]_g=0$ for any $\epsilon\geq 0$. 
\end{proposition}
\begin{proof}
	For $g\in\I^{\neq 0}(\bm{x}_k)\bigcap\mathcal{I}^{\neq 0}(\bm{x}_{k+1})$, by line~\ref{line:half_space_project_start}-\ref{line:half_space_end} in Algorithm~\ref{alg:main.hspg.outline}, it is equivalent to  
	\begin{equation}
		\begin{split}
			\left[\bm{x}_k-\alpha_k\Grad f_{\mathcal{B}_k}(\bm{x}_k)-\alpha_k\lambda \frac{[\bm{x}_k]_g}{\norm{[\bm{x}_k]_g}}\right]_g^\top[\bm{x}_k]_g> \epsilon \norm{[\bm{x}_k]_g}^2,\\
			[\hat{\bm{x}}_{k+1}]_g^\top[\bm{x}_k]_g -\alpha_k\lambda \norm{[\bm{x}_k]_g} > \epsilon \norm{[\bm{x}_k]_g}^2,\\
			[\hat{\bm{x}}_{k+1}]_g^\top[\bm{x}_k]_g > (\alpha_k\lambda+\epsilon\norm{[\bm{x}_k]_g})\norm{[\bm{x}_k]_g}.
		\end{split}
	\end{equation}
	Similarly, $g\in\I^{\neq 0}(\bm{x}_k)\bigcap\mathcal{I}^{0}(\bm{x}_{k+1})$ is equivalent to 
	\begin{equation}\label{eq:xkp1_equals_zero}
		\begin{split}
			\left[\bm{x}_k-\alpha_k\Grad f_{\mathcal{B}_k}(\bm{x}_k)-\alpha_k\lambda \frac{[\bm{x}_k]_g}{\norm{[\bm{x}_k]_g}}\right]_g^\top[\bm{x}_k]_g\leq \epsilon \norm{[\bm{x}_k]_g}^2,\\
			[\hat{\bm{x}}_{k+1}]_g^\top[\bm{x}_k]_g -\alpha_k\lambda \norm{[\bm{x}_k]_g} \leq \epsilon \norm{[\bm{x}_k]_g}^2,\\
			[\hat{\bm{x}}_{k+1}]_g^\top[\bm{x}_k]_g \leq (\alpha_k\lambda+\epsilon\norm{[\bm{x}_k]_g})\norm{[\bm{x}_k]_g}.
		\end{split}    
	\end{equation}
	If $\norm{[\hat{\bm{x}}_{k+1}]_g}\leq \alpha_k\lambda$, then 
	\begin{equation}
		[\hat{\bm{x}}_{k+1}]_g^\top[\bm{x}_k]_g \leq \norm{[\hat{\bm{x}}_{k+1}]_g}\norm{[\bm{x}_k]_g}\leq \alpha_k\lambda \norm{[\bm{x}_k]_g}.
	\end{equation}
	Hence $[\bm{x}_{k+1}]_g=0$ holds for any $\epsilon\geq 0$ by~\eqref{eq:xkp1_equals_zero}, which implies that the projection region of~\proxsg{} and its variance reduction variants, \eg,~\proxsvrg{},~\proxspider{} and~\saga{} are the subsets of~\hspg{}'s.
\end{proof}


\subsection{Convergence Analysis of \halfspacestep{} }\label{appendix.hspg.convergence_analysis}

In this section, we give the convergence result of \halfspacestep{} under the following Assumptions for the properties of the objective function and the global optimal solution $\bm{x}^*$ of \eqref{prob.main}. 


\begin{assumption} \label{assumption:hs-obj-property}
	Assume the following assumptions hold.
	\begin{itemize}
		\item \textbf{(A\ref{assumption:hs-obj-property}-1).} For $i=1,2,\cdots, N$, each $f_i:\mathbb{R}^n\to \mathbb{R}$ is differentiable and bounded below.
		\item \textbf{(A\ref{assumption:hs-obj-property}-2).} For $i=1,2,\cdots, N$, each $f_i: \mathbb{R}^n\to \mathbb{R}$ is $L_i$ smooth. 
		\item \textbf{(A\ref{assumption:hs-obj-property}-3).} $\psi_{\B} = f_{\B} + \lambda r: \mathbb{R}^n \mapsto \mathbb{R}$ has bounded sub-gradient (i.e., $\mathbb{E}[\|\nabla \psi_{\B}(\bm{x})\|^2] \leq M^2$ for some universal constant $M$) over all points $\bm{x} \in \mathbb{R}^n$ with respect to any mini-batch $\B \subseteq [N]$.  
		\item \textbf{(A\ref{assumption:hs-obj-property}-4).} The stochastic gradient $\nabla f_{\mathcal{B}}(\bm{x})$ satisfies $\mathbb{E}_{\mathcal{B}}[\nabla f_{\mathcal{B}}(\bm{x})| \bm{x}] = \nabla f(\bm{x})$ for all $\bm{x} \in \mathbb{R}^n$.
		\item \textbf{(A\ref{assumption:hs-obj-property}-5).} The stochastic gradient $\nabla f_{\mathcal{B}}(\bm{x})$ satisfies $\text{Var}_{\mathcal{B}}[\nabla f_{\mathcal{B}}(\bm{x})| \bm{x}] \leq \sigma^2$ for all $\bm{x} \in \mathbb{R}^n$, where $\sigma^2 > 0$ is a universal constant. 
	\end{itemize}
\end{assumption}
Notice that this Assumption~\ref{assumption:hs-obj-property} is a variant of the Assumption~\ref{assumption:hspg-sufficient-decrease}, to be concise, we set $L$ proposed in Assumption~\ref{assumption:hspg-sufficient-decrease} as $L := \max_{i = 1}^N \{L_i\}$.  

\begin{assumption}
	\label{assumption:stepsize}
	Assume the following assumptions hold. 
	\begin{itemize}
		\item \textbf{(A\ref{assumption:stepsize}-1).} $\sum_{k \geq N_{\mathcal{P}}} \alpha_k = \infty.$
		\item \textbf{(A\ref{assumption:stepsize}-2).} $\sum_{k \geq N_{\mathcal{P}}} \alpha_k^2 < \infty.$
	\end{itemize}
\end{assumption}

\begin{assumption}
	\label{assumption:opt-solution-property} 
	The least and the largest $\ell_2$-norm of non-zero groups in $\bm{x}^*$ are lower and upper bounded by some constants,
	\begin{align}
		0< 2 \delta_1 :=\min_{g\in \mathcal{I}^{\neq 0}(\bm{x}^*)}\norm{[\bm{x}^*]_g} \leq \max_{g\in \mathcal{I}^{\neq 0}(\bm{x}^*)}\norm{[\bm{x}^*]_g} =: 2 \delta_2. 
	\end{align}

\end{assumption}

\begin{theorem} \label{thm:convergence}
	Under Assumptions~(\ref{assumption:hspg-sufficient-decrease}, \ref{assumption:hs-obj-property},  \ref{assumption:stepsize}, \ref{assumption:opt-solution-property}), set 
	\begin{align}
		R \in & ~ \left( 0, ~ \min \left\{\frac{1}{\epsilon} \cdot \left[-(\delta_1+2\epsilon\delta_2)+\sqrt{(\delta_1+2\epsilon\delta_2)^2-4\epsilon^2\delta_2+4\epsilon\delta_1^2}\right], \delta_1 \right\} \right),\\
		\epsilon \in & ~ \left[ 0, ~  \min\left\{\frac{\delta_1^2}{\delta_2}, \frac{2\delta_1-R}{2\delta_2+R}\right\} \right), \\
		\alpha_k \in & ~ \left(0, ~  \min\left\{\frac{2(1-\epsilon)}{L}, \frac{1}{L},\frac{2\delta_1-R-\epsilon(2\delta_2+R)}{M}\right\} \right), ~~~~  \forall k \geq N_{\mathcal{P}}. 
	\end{align}
	If there exists a $K \geq N$ such that 
	\begin{align}
		\|\bm{x}_K - \bm{x}^*\| \leq \frac{R}{2}.
	\end{align}
	Given any $\tau \in (0,1)$, there exists some $ \alpha_{k}= \O(1/(1 + \sqrt{\tau})(k - K))$ and $|\B_{k}|=\O(k - K)$ for all $k \geq K$ such that the sequence $\{\bm{x}_k\}_{k \geq K}$ obtained from the Algorithm~\ref{alg:main.hspg.outline} converges to some stationary point with probability at least $1 - \tau$, i.e., 
	\begin{align}
		\liminf_{k} \mathbb{E}\left[ \norm{\nabla \psi_{\mathcal{B}_k}(\bm{x}_k)}\right]=0 ~~~~ \text{with probability} ~~~~ 1 - \tau. 
	\end{align}
\end{theorem}


\begin{proof}
	\textbf{Proof Sketch.} We split the proof of showing the convergence to some stationary points into two parts. In the first part, we show the convergence holds for all groups in $\tilde{\mathcal{G}}_k$; and in the second part, we show the convergence also holds in $\hat{\mathcal{G}}_k$. 
	
	\textbf{Convergence in $\tilde{\mathcal{G}}_k$ part.} 
	For any $t \in \mathbb{N}_+$, applying Corollary~\ref{corollary:psi_epoch_decrease} yields
	\begin{align}
		& ~ \psi(\bm{x}_{N_{\mathcal{P}}}) - \psi(\bm{x}_{N_{\mathcal{P}} + t}) \\
		= & ~ \sum_{k = N_{\mathcal{P}}}^{N_{\mathcal{P}} + t - 1} \psi(\bm{x}_{k}) - \psi(\bm{x}_{k + 1}) \\
		\geq & ~ \sum_{k = N_{\mathcal{P}}}^{N_{\mathcal{P}} + t - 1} \sum_{g\in\tilde{\mathcal{G}}_k}\left(\alpha_k-\frac{\alpha_k^2L}{2}\right)\mathbb{E}\left[\norm{[\nabla \psi_{\mathcal{B}_k}(\bm{x}_k)]_g}^2\right] + \sum_{k = N_{\mathcal{P}}}^{N_{\mathcal{P}} + t - 1} \left(\frac{1-\epsilon}{\alpha_k}-\frac{L}{2}\right)\sum_{g\in\hat{\G}_k}\norm{[\bm{x}_{k}]_g}^2.
	\end{align}
	Combining the assumption that $\psi$ is bounded below and letting $t \rightarrow \infty$ yield
	\begin{align}
		\underbrace{ \sum_{k = N_{\mathcal{P}}}^{\infty} \sum_{g\in\tilde{\mathcal{G}}_k}\left(\alpha_k-\frac{\alpha_k^2L}{2}\right)\mathbb{E}\left[\norm{[\nabla \psi_{\mathcal{B}_k}(\bm{x}_k)]_g}^2\right] }_{=: T_1} + \underbrace{ \sum_{k = N_{\mathcal{P}}}^{\infty} \left(\frac{1-\epsilon}{\alpha_k}-\frac{L}{2}\right)\sum_{g\in\hat{\G}_k}\norm{[\bm{x}_{k}]_g}^2}_{=: T_2} < \infty. 
	\end{align}
	Given $\alpha_k \in (0, 2(1 - \epsilon) / L)$, we have $T_1 > 0, T_2 > 0$, combining with $T_1 + T_2 < \infty$ implies
	\begin{align}
		& ~ \sum_{k = N_{\mathcal{P}}}^{\infty} \sum_{g\in\tilde{\mathcal{G}}_k}\left(\alpha_k-\frac{\alpha_k^2L}{2}\right)\mathbb{E}\left[\norm{[\nabla \psi_{\mathcal{B}_k}(\bm{x}_k)]_g}^2\right] \\
		= & ~ \sum_{k = N_{\mathcal{P}}}^{\infty} \sum_{g\in\tilde{\mathcal{G}}_k} \alpha_k \mathbb{E}\left[\norm{[\nabla \psi_{\mathcal{B}_k}(\bm{x}_k)]_g}^2\right] - \sum_{k = N_{\mathcal{P}}}^{\infty} \sum_{g\in\tilde{\mathcal{G}}_k} \frac{\alpha_k^2 L }{2} \mathbb{E}\left[\norm{[\nabla \psi_{\mathcal{B}_k}(\bm{x}_k)]_g}^2\right].
	\end{align}
	Based on the boundness of sub-gradient in Assumptions~\ref{assumption:hs-obj-property} and the choice of stepsize in \ref{assumption:stepsize}, we have
	\begin{align}
		\sum_{k = N_{\mathcal{P}}}^{\infty} \sum_{g\in\tilde{\mathcal{G}}_k} \frac{\alpha_k^2 L }{2} \mathbb{E}\left[\norm{[\nabla \psi_{\mathcal{B}_k}(\bm{x}_k)]_g}^2\right] < \infty,
	\end{align}
	which yields 
	\begin{align}
		& ~ \sum_{k = N_{\mathcal{P}}}^{\infty} \sum_{g\in\tilde{\mathcal{G}}_k} \alpha_k \mathbb{E}\left[\norm{[\nabla \psi_{\mathcal{B}_k}(\bm{x}_k)]_g}^2\right] < \infty \\
		\Rightarrow ~&~ \liminf_{k \geq N_{\mathcal{P}}} \sum_{g\in\tilde{\mathcal{G}}_k}  \mathbb{E}\left[\norm{[\nabla \psi_{\mathcal{B}_k}(\bm{x}_k)]_g}^2\right] = 0 \\
		\Rightarrow ~&~ \lim_{k \geq \mathcal{K}} \sum_{g\in\tilde{\mathcal{G}}_k}  \mathbb{E}\left[\norm{[\nabla \psi_{\mathcal{B}_k}(\bm{x}_k)]_g}^2\right] = 0, ~~~ \exists ~ \mathcal{K} \subseteq \{N_{\mathcal{P}}, \ldots \}
		\label{series:sum_convergent_half_space_grad_Psi_subsequence} 
	\end{align}
	
	\textbf{Convergence in $\hat{\mathcal{G}}_k$ part.}
	Under Assumption~\ref{assumption:opt-solution-property}, Lemma~(\ref{lemma:support_cover}, \ref{lemma:x_star_in_polyhedron}, \ref{lemma.project_as_zero_group}) show that if there exists a $K \geq N_{\mathcal{P}}$ such that 
	\begin{align}
		\|\bm{x}_K - \bm{x}^*\| \leq R, 
	\end{align}
	then we have the following results hold
	\begin{align}
		& ~ \I^{\neq 0}(\bm{x}^*) \subseteq \I^{\neq 0}(\bm{x}_K), & \text{non-zero group coverage},\\
		& ~ \bm{x}^* \in \mathcal{S}_K, & \text{correct optimal inclusion $\mathcal{S}_K$},\\
		& ~ \I^{\neq 0}(\bm{x}_{K}) \cap \I^{= 0}(\bm{x}_{K + 1}) \subseteq \I^{= 0}(\bm{x}^*), & \text{correct zero group projection}. 
	\end{align}
	Under Assumption~(\ref{assumption:hs-obj-property}, \ref{assumption:stepsize}, \ref{assumption:opt-solution-property}), Lemma (\ref{lemma:convergence-series}, \ref{lemma:k_plus_1_optimal_dist_non_increase}, \ref{lemma:x_k_in_neghibors}) and Corollary~\ref{corollary:x_star_in_all_polyhedrons} show that: given any $\tau \in (0,1)$, with probability at least $1-\tau$, for any $k \geq K$, $\bm{x}^*$ inhabits $\S_k$. Therefore, for any $k \geq K$, any group $g \in \hat{\mathcal{G}}_k = \I^{\neq 0}(\bm{x}_{k}) \cap \I^{= 0}(\bm{x}_{k + 1})$ will be projected to zero group correctly with probability at least $1-\tau$.
	
	\textbf{Convergence over the whole space.}
	Based on the discussion in $\hat{\mathcal{G}}_k$ part, it is sufficient to focus on the subspace of $\tilde{\mathcal{G}}_k$. Hence,~\eqref{series:sum_convergent_half_space_grad_Psi_subsequence} naturally implies that the sequence $\{\bm{x}_k\}_{k\in\mathcal{K}}$ converges to some stationary point with high probability. By the above, we conclude that 
	\begin{equation}
		\mathbb{P}\left(\liminf_{k} \mathbb{E}\left[ \norm{\nabla \psi_{\mathcal{B}_k}(\bm{x}_k)}\right]=0 \right)\geq 1-\tau.   
	\end{equation}
\end{proof}

\subsubsection{Support Lemma in the Proof of
	Theorem~\ref{thm:convergence}}

The Lemma~\ref{lemma:support_cover} shows that if the optimal distance from the current iterate $\bm{x}_k$ to any local minimizer $\bm{x}^*$ is sufficiently small, then~\hspg{} already covers the supports of $\bm{x}^*$,~\ie, $\I^{\neq 0}(\bm{x}^*)\subseteq \I^{\neq 0}(\bm{x}_k)$. 

\begin{lemma}\label{lemma:support_cover}
	Under Assumption~\ref{assumption:opt-solution-property}, given any $R \leq \delta_1$, for any $k \geq N_{\mathcal{P}}$, if $\norm{\bm{x}_k-\bm{x}^*}\leq R$, then we have $\I^{\neq 0}(\bm{x}^*)\subseteq \I^{\neq 0}(\bm{x}_k)$.    
\end{lemma}
\begin{proof}
	For any $g\in \I^{\neq 0}(\bm{x}^*)$, we have that 
	\begin{equation}
		\begin{split}
			\norm{[\bm{x}^*]_g}-\norm{[\bm{x}_k]_g}&\leq \norm{[\bm{x}_k-\bm{x}^*]_g}\leq \norm{\bm{x}_k-\bm{x}^*}\leq R\leq \delta_1\\
			\norm{[\bm{x}_k]_g}&\geq \norm{[\bm{x}^*]_g}-\delta_1\geq 2\delta_1-\delta_1=\delta_1>0
		\end{split}
	\end{equation}
	Hence $\norm{[\bm{x}_k]_g}\neq 0$, \ie, $g\in \I^{\neq 0}(\bm{x}_k)$. Therefore, $\I^{\neq 0}(\bm{x}^*)\subseteq \I^{\neq 0}(\bm{x}_k)$.
\end{proof}

The Lemma~\ref{lemma:x_star_in_polyhedron} shows that if the distance between the current iterate $\bm{x}_k$ and $\bm{x}^*$, \ie, $\norm{\bm{x}_k-\bm{x}^*}$ is sufficiently small, then $\bm{x}^*$ inhabits the reduced space $\S_k:=\S(\bm{x}_k)$.

\begin{lemma}\label{lemma:x_star_in_polyhedron} 
	Under Assumption~\ref{assumption:opt-solution-property}, for any $k \geq N_{\mathcal{P}}$, given $\epsilon \in [0, \delta_1^2/ \delta_2)$ and 
	\begin{align}
		R \leq R^* := \frac{1}{\epsilon} \cdot \left[-(\delta_1+2\epsilon\delta_2)+\sqrt{(\delta_1+2\epsilon\delta_2)^2-4\epsilon^2\delta_2+4\epsilon\delta_1^2}\right], \label{eq:R-star-defintion}
	\end{align}
	if $\norm{\bm{x}_{k}-\bm{x}^*}\leq R$, we have 
	\begin{equation}
		[\bm{x}_{k}]_g^\top[\bm{x}^*]_g\geq \epsilon\norm{[\bm{x}_k]_g}^2, ~~~ g\in\mathcal{I}^{\neq 0}(\bm{x}^*). 
	\end{equation} 
	Consequently, it implies $\bm{x}^*\in\S_k$ by the definition as~\eqref{def:polytope}.
\end{lemma}
\begin{proof}
	For any $g\in\mathcal{I}^{\neq 0}(\bm{x}^*)$, 
	\begin{equation}
		\begin{split}
			\norm{[\bm{x}_k]_g}\leq \norm{[\bm{x}^*]_g} +R\leq 2\delta_2+R,
		\end{split}
	\end{equation}
	and the $R^*$ defined in \eqref{eq:R-star-defintion} is one of the roots of the quadratic $\epsilon z^2+(4\epsilon\delta_2+2\delta_1)z+4\epsilon\delta_2^2-4\delta_1^2=0$ regarding $z\in \mathbb{R}$. Thus 
	\begin{equation}
		\begin{split}
			[\bm{x}_{k}]_g^\top[\bm{x}^*]_g=&[\bm{x}_{k}-\bm{x}^*+\bm{x}^*]_g^\top [\bm{x}^*]_g\\
			=&[\bm{x}_{k}-\bm{x}^*]_g^\top[\bm{x}^*]_g+\norm{[\bm{x}^*]_g}^2\\
			\geq& \norm{[\bm{x}^*]_g}^2-\norm{[\bm{x}_k-\bm{x}^*]_g}\norm{[\bm{x}^*]_g}\\
			=& \norm{[\bm{x}^*]_g}(\norm{[\bm{x}^*]_g}-\norm{[\bm{x}_k-\bm{x}^*]_g})\\
			\geq & 2\delta_1(2\delta_1-R)\geq \epsilon (2\delta_2+R)^2\\
			\geq &\epsilon\norm{[\bm{x}_k]_g}^2
		\end{split}
	\end{equation}
	holds for any $g\in\mathcal{I}^{\neq 0}(\bm{x}^*)$, where the second last inequality holds because that $2\delta_1(2\delta_1-R)=\epsilon(2\delta_2+R)^2$ as $R=R^*$. Now combing with the definition of $\S_k$ as~\eqref{def:polytope}, we have $\bm{x}^*$ inhabits $\S_k$, which completes the proof.
\end{proof}

The Lemma~\ref{lemma.project_as_zero_group} shows that if $\norm{\bm{x}_k-\bm{x}^*}$ is small enough and the step size is selected properly, every recovery of group sparsity by~\halfspacestep{} can be guaranteed as successful as stated in the following lemma.
\begin{lemma}\label{lemma.project_as_zero_group} 
	Under Assumption~\ref{assumption:opt-solution-property}, for any $k\geq N_\P$, given $\epsilon \in \left[0, \frac{2\delta_1-R}{2\delta_2+R}\right)$, $\alpha_k\in \left(0, \frac{2\delta_1-R-\epsilon(2\delta_2+R)}{M}\right)$ and $R \in (0, \min\{R^*, \delta_1\})$, if $\norm{\bm{x}_{k}-\bm{x}^*}\leq R$, then for any $g\in\hat{\G}_k={\I^{\neq 0}(\bm{x}_k)}\bigcap \I^0(\bm{x}_{k+1})$, we have $g\in\I^0(\bm{x}^*)$.
\end{lemma}
\begin{proof}
	To prove it by contradiction, suppose there exists some $g\in\hat{\G}_k$ such that $g\in \I^{\neq 0}(\bm{x}^*)$. Since $g\in\hat{\G}_k={\I^{\neq 0}(\bm{x}_k)}\bigcap \I^0(\bm{x}_{k+1})$, then the group projection~\eqref{def:proj} is trigerred at $g$ such that
	\begin{equation}\label{eq:hypothesis}
		\begin{split}
			[\tilde{\bm{x}}_{k+1}]_{g}^\top[\bm{x}_k]_{g}&=[\bm{x}_k-\alpha\Grad \psi_{\B_k}(\bm{x}_k)]_{g}^\top[\bm{x}_k]_{g}\\
			&=\norm{[\bm{x}_k]_{g}}^2-\alpha_k [\Grad \psi_{\B_k}(\bm{x}_k)]_{g}^\top [\bm{x}_k]_{g}< \epsilon \norm{[\bm{x}_k]_g}^2.
		\end{split}
	\end{equation}	
	On the other hand, it follows the assumption of this lemma and $g\in\I^{\neq 0}(\bm{x}^*)$ that 
	\begin{equation}
		\norm{[\bm{x}_{k}-\bm{x}^*]_{g}}\leq \norm{\bm{x}_k-\bm{x}^*}\leq R 
	\end{equation}
	Combining the definition of $\delta_1$ and $\delta_2$ in Assumption~\ref{assumption:opt-solution-property}, we have that 
	\begin{equation}
		\begin{split}
			\norm{[\bm{x}_k]_{g}}\geq \norm{[\bm{x}^*]_{g}}-R\geq  2\delta_1 -R\\
			\norm{[\bm{x}_k]_{g}}\leq \norm{[\bm{x}^*]_{g}}+R\leq  2\delta_2 +R\\
		\end{split}
	\end{equation}
	It then follows $0<\alpha_k\leq\frac{2\delta_1-R-\epsilon(2\delta_2+R)}{M}$, where note $2\delta_1-R-\epsilon(2\delta_2+R)>0$ as $R\leq \delta_1$ and $\epsilon<\frac{2\delta_1-R}{2\delta_2+R}$, that 
	\begin{equation}
		\begin{split}
			[\tilde{\bm{x}}_{k+1}]_{g}^\top[\bm{x}_k]_{g}&=\norm{[\bm{x}_k]_{g}}^2-\alpha_k [\Grad \psi_{\B_k}(\bm{x}_k)]_{g}^\top [\bm{x}_k]_{g}\\
			&\geq \norm{[\bm{x}_k]_{g}}^2-\alpha_k\norm{[\Grad \psi_{\B_k}(\bm{x}_k)]_g}\norm{[\bm{x}_k]_g}\\
			&= \norm{[\bm{x}_k]_{g}}(\norm{[\bm{x}_k]_{g}}-\alpha_k\norm{[\Grad \psi_{\B_k}(\bm{x}_k)]_g})\\
			&\geq  \norm{[\bm{x}_k]_{g}}(\norm{[\bm{x}_k]_{g}}-\alpha_kM)\\
			&\geq  \norm{[\bm{x}_k]_{g}}\left[(2\delta_1-R)-\alpha_kM\right]\\
			&\geq \norm{[\bm{x}_k]_{g}}\left[(2\delta_1-R)-\frac{2\delta_1-R-\epsilon(2\delta_2+R)}{M}M\right]\\ 
			&\geq \norm{[\bm{x}_k]_{g}}\left[(2\delta_1-R)-2\delta_1+R+\epsilon(2\delta_2+R)\right]\\ 
			&\geq \epsilon\norm{[\bm{x}_k]_{g}}(2\delta_2+R)\\
			&\geq \epsilon\norm{[\bm{x}_k]_g}^2
		\end{split}
	\end{equation}
	which contradicts with~\eqref{eq:hypothesis}. Hence, we conclude that any $g$ of variables projected to zero, \ie, $g\in\hat{\G}_k={\I^{\neq 0}(\bm{x}_k)}\bigcap \I^0(\bm{x}_{k+1})$ are exactly also the zeros on the optimal solution $\bm{x}^*$, \ie, $g\in\I^{0}(\bm{x}^*)$. 
\end{proof}

We next present that if the iterate of~\halfspacestep{} is close enough to the optimal solution $\bm{x}^*$, then $\bm{x}^*$ inhabits all  reduced spaces constructed by the subsequent iterates of~\halfspacestep{} with high probability. 

To establish this results, we require the following two lemmas (Lemma~\ref{lemma:convergence-series} and Lemma~\ref{lemma:k_plus_1_optimal_dist_non_increase}). The Lemma~\ref{lemma:convergence-series} bounds the accumulated error because of random sampling. Here we introduce the error of gradient estimator on $\I^{\neq 0}(\bm{x})$ for $\psi$ on mini-batch $\B$ as
\begin{equation}\label{def:error_b}
	\bm{e}_{\B}(\bm{x}):=[\Grad \psi_{\B}(\bm{x})-\Grad \psi(\bm{x})]_{\I^{\neq 0}(\bm{x})},
\end{equation}
where by the definition of $r$ in problem~\eqref{prob.main-appendix}, we have $\bm{e}_{\B}(\bm{x})$ also equals to the error of estimation for $\Grad f$, i.e., $\bm{e}_{\B}(\bm{x}) = [\Grad f_{\B}(\bm{x})-\Grad f(\bm{x})]_{\I^{\neq 0}(\bm{x})}.$

\begin{lemma}\label{lemma:convergence-series}
	Under Assumption~\ref{assumption:hs-obj-property}, 
	given any $\theta > 1$, $K\geq N_\P$, let $k:=K+t$ with $t\in\mathbb{Z}_{\geq 0}$, then there exists a sequence of stepsize $\alpha_k = \O(1/(1 + \theta) t)$ and corresponding size of mini-batch $|\B_k|=\O(t)$, such that for any $y_t\in \mathbb{R}^n$,  
	\begin{align*}
		\max_{\{\bm{y}_t\}_{t = 0}^{\infty} \in \mathcal{X}^{\infty}} \sum_{t = 0}^{\infty} \alpha_{k} \|e_{\mathcal{B}_{k}}(\bm{y}_{t})\|_2 \leq \frac{3R^2}{8(4R + 1)}  
	\end{align*}
	holds with probability at least $1 - \frac{1}{\theta^2}$.
\end{lemma}

\begin{proof}
	Define random variable $Y_t := \alpha_{K + t} \|e_{\mathcal{B}_{K + t}}(\bm{y}_{t})\|_2$ for all $t \geq 0$. Since $\{\bm{y}_t\}_{t = 0}^{\infty}$ are arbitrarily chosen, then the random variables $\{Y_t\}_{t = 0}^{\infty}$ are independent. Let $Y := \sum_{t= 0}^{\infty} Y_t$. Using Chebshev's inequality, we obtain
	\begin{align}\label{eq:chevshev_inequality}
		\mathbb{P}\left( Y \geq \mathbb{E}[Y] + \theta \sqrt{\text{Var}[Y]} \right) \leq \mathbb{P}\left( |Y - \mathbb{E}[Y]| \geq \theta \sqrt{\text{Var}[Y]} \right) \leq \frac{1}{\theta^2}. 
	\end{align}
	And based on the Assumption~\ref{assumption:hs-obj-property}, there exists an upper bound $\sigma^2>0$ for the variance of random noise $e_{\B}(\bm{x})$ generated from the one-point mini-batch, \ie, $\mathcal{B}=\{i\}, i = 1,\ldots, N$. Consequently, for each $t \geq 0$, we have $\mathbb{E}[Y_t] \leq \frac{\alpha_{K + t} \sigma}{\sqrt{|\mathcal{B}_{K + t}|}}$ and $\text{Var}[Y_t] \leq \frac{\alpha_{K + t}^2 \sigma^2}{|\mathcal{B}_{K + t}|}$, then combining with~\eqref{eq:chevshev_inequality}, we have
	\begin{align}
		Y &\leq \mathbb{E}[Y] + \theta \sqrt{\text{Var}[Y]} \\
		&\leq \sum_{t = 0}^{\infty} \frac{\alpha_{K + t} \sigma}{\sqrt{|\mathcal{B}_{k + t}|}} + \theta \cdot \sum_{t = 0}^{\infty} \frac{\alpha_{K + t}^2 \sigma^2}{|\mathcal{B}_{K + t}|}\\
		&\leq \sum_{t = 0}^{\infty} \frac{\alpha_{K + t} \sigma}{\sqrt{|\mathcal{B}_{k + t}|}} + \theta \cdot \sum_{t = 0}^{\infty} \frac{\alpha_{K + t} \sigma}{\sqrt{|\mathcal{B}_{K + t}|}} =(1+\theta)\sum_{t = 0}^{\infty} \frac{\alpha_{K + t} \sigma}{\sqrt{|\mathcal{B}_{K + t}|}}
	\end{align}
	holds with probability at least $1 - \frac{1}{\theta^2}$. Here, for the second inequality, we use the property that the equality $\mathbb{E}[\sum_{t = 0}^{\infty} Y_i] = \sum_{t = 0}^{\infty} \mathbb{E}[ Y_i]$ holds whenever $\sum_{t = 0}^{\infty} \mathbb{E}[|Y_i|]$ convergences, see Section 2.1 in \cite{mitzenmacher2005probability}; and for the third inequality, we use $\frac{\alpha_{K + t} \sigma}{\sqrt{|\mathcal{B}_{K + t}|}}\leq 1$ without loss of generality as the common setting of large mini-batch size and small step size.  
	
	Given any $\theta > 1$, there exists some $ \alpha_{k}= \O(1/(1 + \theta) t)$ and $|\B_{k}|=\O(t)$, the above series converges and satisfies that 
	\begin{equation}
		(1+\theta)\sum_{t = 0}^{\infty} \frac{\alpha_{K + t} \sigma}{\sqrt{|\mathcal{B}_{K + t}|}} \leq \frac{3R^2}{8(4R + 1)}
	\end{equation}
	holds. Notice that the above proof holds for any given sequence $\{\bm{y}_t\}_{t = 0}^{\infty} \in \mathcal{X}^{\infty}$, thus  
	\begin{align*}
		\max_{\{\bm{y}_t\}_{t = 0}^{\infty} \in \mathcal{X}^{\infty}} \sum_{t = 0}^{\infty} \alpha_{k} \|e_{\mathcal{B}_{k}}(\bm{y}_{t})\|_2 \leq \frac{3R^2}{8(4R + 1)}
	\end{align*}
	holds with probability at least $1 - \frac{1}{\theta^2}$. 
\end{proof}

The Lemma~\ref{lemma:k_plus_1_optimal_dist_non_increase} draws if previous iterate of~\halfspacestep{} falls into the neighbor of $\bm{x}^*$, then under appropriate step size and mini-batch setting, the current iterate also inhabits the neighbor with high probability.

\begin{lemma}\label{lemma:k_plus_1_optimal_dist_non_increase}
	Under the assumptions of Lemma~\ref{lemma:convergence-series}, suppose $\norm{\bm{x}_{K}-\bm{x}^*}\leq R/2$; for any $\ell$ satisfying $K\leq \ell<K+ t$, $0<\alpha_{\ell}\leq \min\{\frac{1}{L},\frac{2\delta_1-R-\epsilon(2\delta_2+R)}{M}\}$, $|B_\ell|\geq N-\frac{N}{2M}$ and $\norm{\bm{x}_{\ell}-\bm{x}^*}\leq R$ holds, then 
	\begin{equation}
		\norm{\bm{x}_{K+t}-\bm{x}^*}\leq R.
	\end{equation}
	holds with probability at least $1-\frac{1}{\theta^2}$.
\end{lemma}
\begin{proof}
	It follows the assumptions of this lemma, Lemma~\ref{lemma.project_as_zero_group},~\eqref{def:hat_G} and~\eqref{def:tilde_G} that for any $\ell$ satisfying $K\leq \ell<K+ t$
	\begin{equation}
		\norm{[\bm{x}^*]_g}=0,\ \text{for any}\ g\in \hat{\mathcal{G}}_{\ell}.
	\end{equation}
	Hence we have that for $K\leq \ell<K+ t$,
	\begin{equation}\label{eq:optimal_dist_ell}
		\begin{split}
			&\norm{\bm{x}_{\ell+1}-\bm{x}^*}^2\\
			=&\sum_{g\in\tilde{\mathcal{G}}_\ell}\norm{[\bm{x}_{\ell}-\bm{x}^*-\alpha_\ell\Grad \Psi(\bm{x}_{\ell})-\alpha_\ell \bm{e}_{\B_\ell}(\bm{x}_\ell)]_g}^2+\sum_{g\in\hat{\mathcal{G}}_k}\norm{[\bm{x}_{\ell}-\bm{x}^*-\bm{x}_{\ell}]_g}^2\\
			=&\sum_{g\in\tilde{\mathcal{G}}_\ell}\left\{\norm{[\bm{x}_{\ell}-\bm{x}^*]_g}^2-2\alpha_\ell[\bm{x}_{\ell}-\bm{x}^*]_g^\top[\Grad \Psi(\bm{x}_{\ell})+\bm{e}_{\B_\ell}(\bm{x}_\ell)]_g+\alpha_\ell^2\norm{[\Grad \Psi(\bm{x}_{\ell})+\bm{e}_{\B_\ell}(\bm{x}_\ell)]_g}^2\right\}+\sum_{g\in\hat{\G}_\ell}\norm{[\bm{x}^*]_g}^2\\
			=&\sum_{g\in\tilde{\mathcal{G}}_\ell}\left\{\norm{[\bm{x}_{\ell}-\bm{x}^*]_g}^2-2\alpha_\ell[\bm{x}_{\ell}-\bm{x}^*]_g^\top[\Grad \Psi(\bm{x}_{\ell})]_g-2\alpha_\ell[\bm{x}_{\ell}-\bm{x}^*]_g^\top[\bm{e}_{\B_\ell}(\bm{x}_\ell)]_g+\alpha_\ell^2\norm{[\Grad \Psi(\bm{x}_{\ell})+\bm{e}_{\B_\ell}(\bm{x}_\ell)]_g}^2\right\}\\
			\leq&\sum_{g\in\tilde{\G}_\ell}\norm{[\bm{x}_{\ell}-\bm{x}^*]_g}^2-\norm{[\Grad \Psi(\bm{x}_{\ell})]_g}^2\left(2\frac{\alpha_\ell}{L}-\alpha_\ell^2\right)-2\alpha_\ell[\bm{x}_{\ell}-\bm{x}^*]_g^\top[\bm{e}_{\B_\ell}(\bm{x}_\ell)]_g+\alpha_\ell^2\norm{[\bm{e}_{\B_\ell}(\bm{x}_\ell)]_g}^2\\
			&+2\alpha_\ell^2[\Grad \Psi(\bm{x}_\ell)]_g^\top[\bm{e}_{\B_\ell}(\bm{x}_\ell)]_g\\
			\leq& \sum_{g\in\tilde{\G}_\ell}\norm{[\bm{x}_{\ell}-\bm{x}^*]_g}^2-\norm{[\Grad \Psi(\bm{x}_{\ell})]_g}^2\left(2\frac{\alpha_\ell}{L}-\alpha_\ell^2\right) +2\alpha_\ell\norm{[\bm{x}_{\ell}-\bm{x}^*]_g}\norm{[\bm{e}_{\B_\ell}(\bm{x}_\ell)]_g}+\alpha_\ell^2\norm{[\bm{e}_{\B_\ell}(\bm{x}_\ell)]_g}^2\\
			&+2\alpha_\ell^2\norm{[\Grad \Psi(\bm{x}_\ell)]_g}\norm{[\bm{e}_{\B_\ell}(\bm{x}_\ell)]_g}\\
			\leq& \sum_{g\in\tilde{\G}_\ell}\norm{[\bm{x}_{\ell}-\bm{x}^*]_g}^2-\norm{[\Grad \Psi(\bm{x}_{\ell})]_g}^2\left(2\frac{\alpha_\ell}{L}-\alpha_\ell^2\right) +(2\alpha_\ell+2\alpha_\ell^2L)\norm{[\bm{x}_{k}-\bm{x}^*]_g}\norm{[\bm{e}_{\B_\ell}(\bm{x}_\ell)]_g}+\alpha_\ell^2\norm{[\bm{e}_{\B_\ell}(\bm{x}_\ell)]_g}^2\\
			\leq& \sum_{g\in\tilde{\G}_\ell}\left\{\norm{[\bm{x}_{\ell}-\bm{x}^*]_g}^2-\norm{[\Grad \Psi(\bm{x}_{\ell})]_g}^2\left(2\frac{\alpha_\ell}{L}-\alpha_\ell^2\right)\right\} +(2\alpha_\ell+2\alpha_\ell^2L)\norm{\bm{x}_{\ell}-\bm{x}^*}\norm{\bm{e}_{\B_\ell}(\bm{x}_\ell)}+\alpha_\ell^2\norm{\bm{e}_{\B_\ell}(\bm{x}_\ell)}^2
		\end{split}
	\end{equation}	
	
	On the other hand, by the definition of $\bm{e}_{\B}(\bm{x})$ as~\eqref{def:error_b}, we have that 
	\begin{equation}\label{eq:error_eq}
		\begin{split}
			\bm{e}_{\B}(\bm{x})=&[\Grad \Psi_{\B}(\bm{x})-\Grad \Psi(\bm{x})]_{\I^{\neq 0}(\bm{x})}=[\Grad f_{\B}(\bm{x})-\Grad f(\bm{x})]_{\I^{\neq 0}(\bm{x})}\\
			=&\frac{1}{|\B|}\sum_{j\in \B}[\Grad f_{j}(\bm{x})]_{\I^{\neq 0}(\bm{x})}-\frac{1}{N}\sum_{i=1}^N [\Grad f_i(\bm{x})]_{\I^{\neq 0}(\bm{x})}\\
			=&\frac{1}{N}\sum_{j\in\B}\left[\frac{N}{|\B|}[\Grad f_{j}(\bm{x})]_{\I^{\neq 0}(\bm{x})}-[\Grad f_j(\bm{x})]_{\I^{\neq 0}(\bm{x})}\right]-\frac{1}{N}\sum_{\substack{i=1\\ i\notin \B}}^N[\Grad f_i(\bm{x})]_{\I^{\neq 0}(\bm{x})}\\
			=&\frac{1}{N}\sum_{j\in\B}\left[\frac{N-|\B|}{|\B|}[\Grad f_{j}(\bm{x})]_{\I^{\neq 0}(\bm{x})}\right]-\frac{1}{N}\sum_{\substack{i=1\\ i\notin \B}}^N[\Grad f_i(\bm{x})]_{\I^{\neq 0}(\bm{x})}\\
		\end{split}
	\end{equation}
	Thus taking the norm on both side of~\eqref{eq:error_eq} and using triangle inequality results in the following:
	\begin{equation}\label{eq:bound_error}
		\begin{split}
			\norm{\bm{e}_\B(\bm{x})}&\leq \frac{1}{N}\sum_{j\in\B}\left[\frac{N-|\B|}{|\B|}\norm{[\Grad f_{j}(\bm{x})]_{\I^{\neq 0}(\bm{x})}}\right]+\frac{1}{N}\sum_{\substack{i=1\\ i\notin \B}}^N\norm{[\Grad f_i(\bm{x})]_{\I^{\neq 0}(\bm{x})}}\\
			&\leq \frac{1}{N} \frac{N-|\B|}{|\B|} |\B_k| M + \frac{1}{N} (N-|\B|)M\leq \frac{2(N-|\B|)M}{N}.
		\end{split}
	\end{equation}

	Since $\alpha_{\ell}\leq 1$, and $|B_\ell|\geq N-\frac{N}{2M}$ hence $\alpha_{\ell}\norm{\bm{e}_{\B_{\ell}}(\bm{x}_{\ell})}\leq 1$. Then combining with $\alpha_\ell\leq 1/L$,~\eqref{eq:optimal_dist_ell} can be further simplified as 
	\begin{equation}\label{eq:x_kp1_optimal_dist_2}
		\begin{split}
			&\norm{\bm{x}_{\ell+1}-\bm{x}^*}^2\\
			\leq & \sum_{g\in\tilde{\G}_\ell}\left\{\norm{[\bm{x}_{\ell}-\bm{x}^*]_g}^2-\norm{[\Grad \Psi(\bm{x}_{\ell})]_g}^2\left(2\frac{\alpha_\ell}{L}-\alpha_\ell^2\right)\right\} +(2\alpha_\ell+2\alpha_\ell^2L)\norm{\bm{x}_{\ell}-\bm{x}^*}\norm{\bm{e}_{\B_\ell}(\bm{x}_\ell)}+\alpha_\ell^2\norm{\bm{e}_{\B_\ell}(\bm{x}_\ell)}^2\\
			\leq & \sum_{g\in\tilde{\G}_\ell}\left\{\norm{[\bm{x}_{\ell}-\bm{x}^*]_g}^2-\frac{1}{L^2}\norm{[\Grad \Psi(\bm{x}_{\ell})]_g}^2\right\}+4\alpha_\ell\norm{\bm{x}_{\ell}-\bm{x}^*}\norm{\bm{e}_{\B_\ell}(\bm{x}_\ell)}+\alpha_\ell^2\norm{\bm{e}_{\B_\ell}(\bm{x}_\ell)}^2\\
			\leq &\norm{\bm{x}_\ell-\bm{x}^*}^2+4\alpha_\ell\norm{\bm{x}_{\ell}-\bm{x}^*}\norm{\bm{e}_{\B_\ell}(\bm{x}_\ell)}+\alpha_\ell\norm{\bm{e}_{\B_\ell}(\bm{x}_\ell)}
		\end{split}
	\end{equation}
	Following from the assumption that $\norm{\bm{x}_{\ell}-\bm{x}^*}\leq R$, then~\eqref{eq:x_kp1_optimal_dist_2} can be further simplified as 
	\begin{equation}\label{eq:x_kp1_optimal_dist_3}
		\begin{split}
			\norm{\bm{x}_{\ell+1}-\bm{x}^*}^2\leq & \norm{\bm{x}_\ell-\bm{x}^*}^2+4\alpha_\ell R\norm{\bm{e}_{\B_\ell}(\bm{x}_\ell)}+\alpha_k\norm{\bm{e}_{\B_\ell}(\bm{x}_\ell)}\\
			\leq & \norm{\bm{x}_\ell-\bm{x}^*}^2+(4R+1)\alpha_\ell\norm{\bm{e}_{\B_\ell}(\bm{x}_\ell)}
		\end{split}    
	\end{equation}
	Summing the the both side of~\eqref{eq:x_kp1_optimal_dist_3} from $\ell=K$ to $\ell=K+t-1$ results in  
	\begin{equation}
		\begin{split}
			\norm{\bm{x}_{K+t}-\bm{x}^*}^2\leq \norm{\bm{x}_K-\bm{x}^*}^2+(4R+1)\sum_{\ell=K}^{K+t-1}\alpha_{\ell}\norm{\bm{e}_{\B_{\ell}}(\bm{x}_{\ell})}\\
		\end{split}
	\end{equation}
	It follows Lemma~\ref{lemma:convergence-series} that the followng holds with probability at least $1-\frac{1}{\theta^2}$,
	\begin{equation}
		\sum_{\ell = K}^{\infty} \alpha_{\ell} \|\bm{e}_{\mathcal{B}_{\ell}}(\bm{x}_{\ell})\| \leq \frac{3R^2}{4(4R + 1)}.
	\end{equation}
	Thus we have that 
	\begin{equation}
		\begin{split}
			\norm{\bm{x}_{K+t}-\bm{x}^*}^2&\leq \norm{\bm{x}_K-\bm{x}^*}^2+\left(4R+1\right)\sum_{\ell=K}^{K+t-1}\alpha_{\ell}\norm{\bm{e}_{\B_{\ell}}(\bm{x}_{\ell})}\\
			&\leq \norm{\bm{x}_K-\bm{x}^*}^2+\left(4R+1\right)\sum_{\ell = K}^{\infty} \alpha_{\ell} \|\bm{e}_{\mathcal{B}_{\ell}}(\bm{x}_{\ell})\|\\
			&\leq \frac{R^2}{4}+(4R+1)\frac{3R^2}{4(4R+1)}\leq \frac{R^2}{4}+\frac{3R^2}{4}\leq R^2,
		\end{split}
	\end{equation}
	holds with probability at least $1-\frac{1}{\theta^2}$, which completes the proof. 
\end{proof}

Based on the above lemmas, the Lemma~\ref{lemma:x_k_in_neghibors} shows if initial iterate of~\halfspacestep{} locates closely enough to $\bm{x}^*$, step size $\alpha_k$ polynomially decreases, and mini-batch size $\B_k$ polynomially increases, then $\bm{x}^*$ inhabits all subsequent  reduced space $\{\S_k\}_{k=K}^{\infty}$ constructed in~\halfspacestep{} with high probability. 

\begin{lemma}\label{lemma:x_k_in_neghibors}
	If $\norm{\bm{x}_{K}-\bm{x}^*}\leq \frac{R}{2}$, $K\geq N_\P$, $k=K+t$, $t\in\mathbb{Z}^+$, $0<\alpha_k=\O(1/(\sqrt{N}t))\leq \min\{\frac{2(1-\epsilon)}{L}, \frac{1}{L},\frac{2\delta_1-R-\epsilon(2\delta_2+R)}{M}\} $ and $|\B_k|=\O(t)\geq N-\frac{N}{2M}$. Then for any constant $\tau\in (0,1)$, $\norm{\bm{x}_k-\bm{x}^*}\leq R$ with probability at least $1-\tau$ for any $k\geq K$. 
\end{lemma}
\begin{proof}
	It follows Lemma~\ref{lemma:x_star_in_polyhedron} and the assumption of this lemma that $\bm{x}^*\in\S_K$. Moreover, it follows the assumptions of Lemma~(\ref{lemma:convergence-series}, \ref{lemma:k_plus_1_optimal_dist_non_increase}, \ref{lemma:x_k_in_neghibors}), the definition of finite-sum $f(\bm{x})$ in \eqref{prob.main-appendix}, and the bound of error as~\eqref{eq:bound_error} that
	\begin{equation}
		\mathbb{P}(\{\bm{x}_k\}_{k=K}^{\infty}\in \{x: \norm{\bm{x}-\bm{x}^*}\leq R\}^{\infty})\geq \left(1-\frac{1}{\theta^2}\right)^{\O(N-K)}\geq 1-\tau,    
	\end{equation}
	where the last two inequalities comes from that the error vanishing to zero as $|\B_k|$ reaches the upper bound $N$, and $\theta$ is sufficiently large depending on $\tau$ and $\O(N-K)$. 
\end{proof}

\begin{corollary}\label{corollary:x_star_in_all_polyhedrons}
	Lemma~\ref{lemma:x_k_in_neghibors} further implies $\bm{x}^*$ inhabits all subsequent $\S_k$, i.e., $\bm{x}^*\in \S_{k}$  for any $k\geq K$.
\end{corollary}

\subsection{The Initialization Stage} \label{appendix.hspg.initial_sage}

In previous parts, we show that the \halfspacestep{} guarantees to converge to the optimal solution, and ensures to recover the no-zero groups of the optimal solution under some assumptions with a ``close-enough'' initialization point $\bm{x}_{N_{\mathcal{P}}}$. To complete the story, in this part, we show that the iterate obtained from the \textit{Subgradient Descent Update} in Algorithm~\ref{alg:main.hspg.outline} satisfies the ``close-enough'' condition with high probability. Remark here that the proximal methods, such as~\proxsg{}, \proxsvrg{} and~\saga{}, may also serve in the initialization stage. However, for the general regularization $r(\bm{x})$, they may not have closed-form solution for the corresponding inherent subproblems, implying non-explicit update mechanism to the next iterate. Hence, people may have to inconveniently approximate the solutions of proximal operator by other techniques, whereas the sub-gradient method does not have these drawbacks. Therefore, for the generality of~\hspg{}, we select the sub-gradient method in the Initialization Stage by default. 

\subsubsection{Convergence Analysis of Initialization Stage}
In this part, we show that the ``close enough'' condition 
\begin{align}
	\norm{\bm{x}_k-\bm{x}^*}\leq \frac{R}{2}
\end{align}
proposed in Theorem~\ref{thm:convergence} can be achieved via the Initialization Stage (\textit{Subgradient Descent Update}) in Algorithm~\ref{alg:main.hspg.outline} under the Assumption~\ref{assumption:init-convergence}. 
\begin{assumption} \label{assumption:init-convergence}
	Assume the following assumptions hold. 
	\begin{itemize}
		\item \textbf{(A\ref{assumption:init-convergence}-1).} $f: \mathbb{R}^n \mapsto \mathbb{R}$ is differentiable and $\mu$-strongly convex. $r: \mathbb{R}^n \mapsto \mathbb{R}$ is convex. 
		\item \textbf{(A\ref{assumption:init-convergence}-2).} There exists an universal constant $M$ such that the stochastic gradient $\nabla f_{\mathcal{B}}(\bm{x})$ satisfies $\|\nabla f_{\mathcal{B}}(\bm{x})\|_2 \leq M$ for all $\bm{x} \in \mathbb{R}^d$ and mini-batch $\mathcal{B}$.  
		\item \textbf{(A\ref{assumption:init-convergence}-3).} The stochastic gradient $\nabla f_{\mathcal{B}}(\bm{x})$ satisfies $\mathbb{E}_{\mathcal{B}}[\nabla f_{\mathcal{B}}(\bm{x})| \bm{x}] = \nabla f(\bm{x})$ for all $\bm{x} \in \mathbb{R}^n$. 
	\end{itemize}
\end{assumption}

\begin{proposition} \label{prop:init-convergence}
	Under Assumption~\ref{assumption:init-convergence}, for any $R > 0$, any $\tau \in (0,1)$, set 
	\begin{align}
		& ~ N = \left\lceil \log \left( \frac{\tau R}{4 \|\bm{x}_0 - \bm{x}^*\|_2^2} \right) \bigg/ \log \left(1 - \frac{\tau R}{4 M} \right) \right\rceil, \\
		& ~ \alpha_0 = \alpha_1 = \ldots = \alpha_{N_{\mathcal{P}} - 1} = \frac{\tau \mu R}{4 M^2},
	\end{align}
	where $R$ based on the setting of Theorem~\ref{thm:convergence}.  We have the Algorithm~\ref{alg:main.outline} (Subgradient Descent Update) returns a solution $\bm{x}_{N_{\mathcal{P}}}$ that satisfies $\|\bm{x}_{N_{\mathcal{P}}} - \bm{x}^*\|_2 \leq R/2$ with probability $1 - \tau$.  
\end{proposition}
\begin{proof}
	Let $\bm{x}^*$ be the global optimal solution of \eqref{prob.main}. Let $\nabla \psi(\bm{x}) = \nabla f(\bm{x}) + \lambda \zeta(\bm{x})$ and $\nabla \psi_{\B}(\bm{x}) = \nabla f_{\B}(\bm{x}) + \lambda \zeta(\bm{x})$ given any point $\bm{x} \in \mathbb{R}^n$ and mini-batch $\B$. Consider
	\begin{align}
		\|\bm{x}_{k + 1} - \bm{x}^* \|^2_2 = & ~ \|\bm{x}_{k} - \alpha_k \nabla \psi_{\mathcal{B}_k} (\bm{x}_k) - \bm{x}^* \|^2_2 \\
		= & ~ \|\bm{x}_k - \bm{x}^*\|_2^2 - 2 \alpha_k \langle  \nabla \psi_{\mathcal{B}_k} (\bm{x}_k), \bm{x}_k - \bm{x}^* \rangle + \|\alpha_k \nabla \psi_{\mathcal{B}_k} (\bm{x}_k)\|_2^2. 
	\end{align}
	Due to (A1) in Assumption~\ref{assumption:init-convergence}, the $\mu$-strongly convexity of $f$ and the convexity of $r$ yields   
	\begin{align}
		\psi(\bm{x}^*) \geq \psi(\bm{x}_k) + \langle \nabla \psi (\bm{x}_k), \bm{x}^* - \bm{x}_k \rangle + \frac{\mu}{2} \| \bm{x}_k - \bm{x}^* \|_2^2.  
	\end{align}
	Thus
	\begin{align}
		& ~ \|\bm{x}_{k + 1} - \bm{x}^* \|^2_2 \\
		= & ~ \|\bm{x}_k - \bm{x}^*\|_2^2 - 2 \alpha_k \langle  \nabla \psi_{\mathcal{B}_k} (\bm{x}_k), \bm{x}_k - \bm{x}^* \rangle + \|\alpha_k \nabla \psi_{\mathcal{B}_k} (\bm{x}_k)\|_2^2 \\
		= & ~ \|\bm{x}_k - \bm{x}^*\|_2^2 + 2 \alpha_k \langle \nabla \psi_{\mathcal{B}_k} (\bm{x}_k), \bm{x}^* - \bm{x}_k  \rangle + \|\alpha_k \nabla \psi_{\mathcal{B}_k} (\bm{x}_k)\|_2^2 \\
		= & ~ \|\bm{x}_k - \bm{x}^*\|_2^2 + 2 \alpha_k \langle \nabla \psi (\bm{x}_k) - \nabla \psi (\bm{x}_k) + \nabla \psi_{\mathcal{B}_k} (\bm{x}_k), \bm{x}^* - \bm{x}_k  \rangle + \|\alpha_k \nabla \psi_{\mathcal{B}_k} (\bm{x}_k)\|_2^2 \\
		\leq & ~ \|\bm{x}_k - \bm{x}^*\|_2^2 + 2 \alpha_k 
		\left( \psi(\bm{x}^*)  - \psi(\bm{x}_k) - \frac{\mu}{2} \|\bm{x}_k - \bm{x}^*\|_2^2 \right) \\
		& ~ + 2 \alpha_k \langle \nabla \psi_{\mathcal{B}_k} (\bm{x}_k) - \nabla \psi (\bm{x}_k) , \bm{x}^* - \bm{x}_k  \rangle + \|\alpha_k \nabla \psi_{\mathcal{B}_k} (\bm{x}_k)\|_2^2 \\
		\leq & ~ (1 - \alpha_k \mu) \|\bm{x}_k - \bm{x}^*\|_2^2 - 2 \alpha_k (\psi(\bm{x}_k) - \psi(\bm{x}^*)) + \alpha_k^2 \|\nabla \psi(\bm{x}_k)\|_2^2 \\
		& ~ + 2 \alpha_k \langle \nabla \psi_{\mathcal{B}_k} (\bm{x}_k) - \nabla \psi (\bm{x}_k) , \bm{x}^* - \bm{x}_k \rangle \\
		\leq & ~ (1 - \alpha_k \mu)  \|\bm{x}_k - \bm{x}^*\|_2^2 + \alpha_k^2 M^2 + 2 \alpha_k \langle \nabla \psi_{\mathcal{B}_k} (\bm{x}_k) - \nabla \psi (\bm{x}_k) , \bm{x}^* - \bm{x}_k \rangle.
	\end{align}
	Given $\bm{x}_k$, due to (A\ref{assumption:init-convergence}-2) in Assumption~\ref{assumption:init-convergence}, taking expectation over $\mathcal{B}_k$ yields
	\begin{align}
		\mathbb{E}_{\mathcal{B}_k} [\|\bm{x}_{k + 1} - \bm{x}^* \|^2_2 | \bm{x}_k] \leq & ~ (1 - \alpha_k \mu)  \|\bm{x}_k - \bm{x}^*\|_2^2 + \alpha_k^2 M^2, 
	\end{align}
	where the above inequality holds by (A\ref{assumption:init-convergence}-3) in Assumption~\ref{assumption:init-convergence}
	\begin{align}
		\mathbb{E}_{\mathcal{B}_k}[\langle \nabla \psi_{\mathcal{B}_k} (\bm{x}_k) - \nabla \psi (\bm{x}_k) , \bm{x}^* - \bm{x}_k \rangle | \bm{x}_k ] = 0 .
	\end{align}
	For any $k \in \mathbb{N}_+$, any constant $c > 0$, and initial point $\bm{x}_0$, setting $\alpha_k = \frac{\mu}{c M^2}$, apply above inequality recursively yields
	\begin{align}
		\mathbb{E}_{\mathcal{H}} \left[ \|\bm{x}_k - \bm{x}^*\|_2^2 \right] \leq \left(1 - \frac{1}{cM^2} \right)^k \|\bm{x}_0 - \bm{x}^*\|_2^2 + \frac{1}{c}, 
	\end{align}
	where $\mathcal{H} = \{\mathcal{B}_0, \ldots, \mathcal{B}_{k - 1}\}$ denotes the whole history until step $k$. \\
	
	\textbf{Non-asymptotic bounds.} Combine above together, given any $R /2 > 0$, for any $\tau \in (0,1)$, set 
	\begin{align}
		& ~ N = \left\lceil \log \left( \frac{\tau R}{4 \|\bm{x}_0 - \bm{x}^*\|_2^2} \right) \bigg/ \log \left(1 - \frac{\tau R}{4 M} \right) \right\rceil, \\
		& ~ \alpha_0 = \alpha_1 = \ldots = \alpha_{N_{\mathcal{P}} - 1} = \frac{\tau \mu R}{4 M^2},
	\end{align}
	by Markov inequality, we have 
	\begin{align}
		\|\bm{x}_k - \bm{x}^*\|_2 \leq R  /2
	\end{align}
	holds with probability $1 - \tau$. 
\end{proof}


\section{Extensive Numerical Experiments}\label{appendix.hspg.extensive_experiment}

In this Appendix, we include extensive numerical experiments in the view of optimization to demonstrate the superiority of~\hspg{} to other classical proximal methods on the sparsity exploration and the competitiveness on objective convergence in both convex and nonconvex settings. Particularly, in Appendix~\ref{appendix.hspg.experiment.convex}, we provide convex experiments to  \textit{(i)} demonstrate the validness of group sparsity identification of HSPG; \textit{(ii)} present comprehensive comparison to Prox-SG, RDA and Prox-SVRG on benchmark convex problems. In Appendix~\ref{appendix.hspg.experiment.nonconvex}, we show additional nonconvex experiments to reveal the superiority of~\hspg{} to competitors on group sparsity exploration.

\subsection{Convex Experiments}\label{appendix.hspg.experiment.convex}

\paragraph{Linear Regression on Synthetic Data}

We numerically validate the proposed HSPG on group sparsity identification by linear regression problems with $\ell_1/\ell_2$ regularizations using synthetic data. Consider a data matrix $A\in\mathbb{R}^{N\times n}$ consisting of $N$ instances and the target variable $\bm{y}\in\mathbb{R}^N$, we are interested in the following problem:
\begin{equation}\label{eq:lr}
	\minimize{\bm{x}\in\mathbb{R}^n}\ \frac{1}{2N}\|A\bm{x}-\bm{y}\|^2+ \lambda \sum_{g\in \G}\norm{[\bm{x}]_g}.
\end{equation}
Our goal is to empirically show that HSPG is able to identify the ground truth zero groups with synthetic data.
We conduct the experiments as follows: \textit{(i)} generate the data matrix $A$ whose elements are uniformly distributed among $[-1, 1]$; \textit{(ii)} generate a vector $\bm{x}^*$ working as the ground truth solution, where the elements are uniformly distributed among $[-1, 1]$ and the coordinates are equally divided into 10 groups ($|\G|=10$); \textit{(iii)} randomly set a number of groups of $\bm{x}^*$ to be 0 according to a pre-specified group sparsity ratio; \textit{(iv)} compute the target variable $\bm{y}=A\bm{x}^*$; (v) solve the above problem \eqref{eq:lr} for $\bm{x}$ with $A$ and $\bm{y}$ only, and then evaluate the Intersection over Union (IoU) with respect to the identities of the zero groups between the computed solution estimate $\hat{\bm{x}}$ by HSPG and the ground truth $\bm{x}^*$.

We test HSPG on \eqref{eq:lr} under different problem settings. For a slim matrix $A$ where $N\ge n$, we test with various group sparsity ratios among $\{0.1,0.3,0.5,0.7,0.9\}$, and for a fat matrix $A$ where $N<n$, we only test with a certain group sparsity value since a recovery of $\bm{x}^*$ requires that the number of non-zero elements in $\bm{x}^*$ is bounded by $N$. Throughout the experiments, we set $\lambda$ to be $100/N$, the mini-batch size $|\mathcal{B}|$ to be 64, step size $\alpha_k$ to be 0.1 (constant), and fine-tune $\epsilon$ per problem. Based on a similar statistical test on objective function stationarity~\cite{zhang2020statistical}, we switch to \halfspacestep{} roughly after 30 epoches. Table~\ref{tb:lr} shows that under each setting, the proposed HSPG correctly identifies the  groups of zeros as indicated by $\textrm{IoU}(\hat{\bm{x}},\bm{x}^*)=1.0$, which is a strong evidence to show the correctness of group sparsity identification of HSPG. 

\begin{table}[h]
	\centering
	\caption{Linear regression problem settings and IoU of the recovered solutions by HSPG.}
	\label{tb:lr}
	\begin{tabular}{c|cccc}
		\hline
		& \quad $N$\quad     &  \quad $n$  \quad   &  \quad Group sparsity ratio of $\bm{x}^*$    \quad             &  \quad IoU($\hat x,x^*$)  \quad \\ \hline
		\multirow{4}{*}{\begin{tabular}[c]{@{}l@{}} Slim $A$ \\  \end{tabular}} & \quad 10000\quad  &  \quad1000 \quad &  \quad\{0.1, 0.3, 0.5, 0.7, 0.9\}  \quad&  \quad1.0  \quad\\ 
		& \quad 10000\quad  &  \quad2000 \quad &  \quad\{0.1, 0.3, 0.5, 0.7, 0.9\} \quad &  \quad1.0 \quad \\ 
		& \quad 10000\quad  &  \quad3000 \quad &  \quad\{0.1, 0.3, 0.5, 0.7, 0.9\} \quad &  \quad1.0  \quad\\ 
		& \quad 10000\quad  &  \quad4000 \quad &  \quad\{0.1, 0.3, 0.5, 0.7, 0.9\}  \quad&  \quad1.0 \quad \\ \hline
		\multirow{4}{*}{\begin{tabular}[c]{@{}l@{}}Fat $A$\\ \end{tabular}}      &  \quad200  \quad  &  \quad1000 \quad&  \quad0.9    \quad                        & \quad 1.0 \quad \\  
		&  \quad300  \quad  &  \quad1000  \quad& \quad 0.8  \quad                          &  \quad1.0  \quad\\ 
		&  \quad400  \quad  &  \quad1000  \quad&  \quad0.7  \quad                          &  \quad1.0  \quad\\ 
		&  \quad500 \quad   &  \quad1000 \quad & \quad 0.6  \quad                          &  \quad1.0  \quad\\ \hline
	\end{tabular}
\end{table}

\paragraph{Logistic Regression} 
We then focus on the benchmark convex logistic regression problem with the mixed $\ell_1/\ell_2$-regularization given $N$ examples $(\bm{d}_1, l_1), \cdots, (\bm{d}_N, l_N)$ where $\bm{d}_i\in \mathbb{R}^n$ and $l_i \in \{-1, 1\}$ with the form\vspace{-0.1cm}
\begin{equation}\label{def:minimize_logistic_l1}
	\small
	\minimize{(\bm{x}; b)\in \R^{n+1}}\ \frac{1}{N}\sum_{i=1}^N \log(1 + e^{-l_i (\bm{x}^T \bm{d}_i +b)}) + \lambda \sum_{g\in \G}\norm{[\bm{x}]_g},
\end{equation}
for binary classification with a bias $b\in\mathbb{R}$. We set the regularization parameter $\lambda$ as $100/N$ throughout the experiments since it yields high sparse solutions and low object value $f$’s, equally decompose the variables into 10 groups to form $\mathcal{G}$, and  test~problem~\eqref{def:minimize_logistic_l1} on 8 standard publicly available large-scale datasets from LIBSVM repository~\cite{chang2011libsvm}
as summarized in Table~\ref{table:datasets}. All convex experiments are conducted on a 64-bit operating system with an Intel(R) Core(TM) i7-7700K CPU $@$ 4.20 GHz and 32 GB random-access memory.

\begin{figure}
	\centering
	\includegraphics[width=0.7\textwidth]{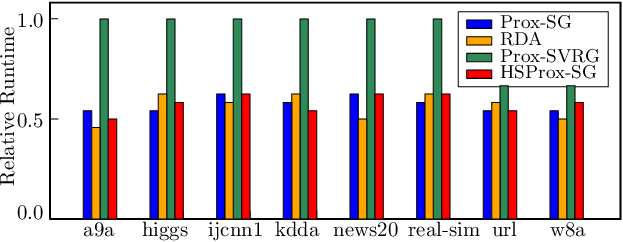}
	\caption{Relative runtime.}
	\label{figure:runtime_convex}
\end{figure}

We run the solvers with a maximum number of epochs as $60$ following~\cite{chen2020orthant}. 
The mini-batch size $|\mathcal{B}|$ is set to be $\min\{256, \lceil{0.01N\rceil}\}$ similarly to~\cite{yang2019stochastic}. The step size $\alpha_k$ setting follows~[Section 4]\cite{xiao2014proximal}. Particularly, we first compute a Lipschitz constant $L$ as $\max_{i}\norm{\bm{d}_i}^2/4$, then fine tune and select constant $\alpha_k\equiv\alpha=1/L$ to~\proxsg{} and~\proxsvrg{} since it exhibits the best results. For~\rda{}, the step size parameter $\gamma$ is fined tuned as the one with the best performance among all powers of $10$. For~\hspg{}, we set $\alpha_k$ as the same as~\proxsg{} and \proxsvrg{} in practice.
We select two $\epsilon$'s as $0$ and $0.8$. The final objective value $\psi$ and group sparsity in the solutions are reported in  Table~\ref{table:object_Psi_value_convex}-\ref{table:group_sparsity_convex}, where we mark the best values as bold to facilitate the comparison. Furthermore, Figure~\ref{figure:runtime_convex} plots the relative runtime of these solvers for each dataset, scaled by the runtime of the most time-consuming solver. 

Table~\ref{table:group_sparsity_convex} shows that our~\hspg{} is definitely the best solver on exploring the group sparsity of the solutions. In fact,~\hspg{} under $\epsilon=0.8$ performs all the best except \textit{ijcnn1}.~\proxsvrg{} is the second best solver on group sparsity exploration, which demonstrates that the variance reduction techniques works well in convex setting to promote sparsity, but not in non-convex settings. ~\hspg{} under $\epsilon=0$ performs much better than~\proxsg{} which matches the better sparsity recovery property of~\hspg{} even under $\epsilon$ as $0$. 
Moreover, as shown in Table~\ref{table:object_Psi_value_convex}, we observe that all solvers perform quite competitively in terms of final objective values (round up to 3 decimals) except~\rda{}, which demonstrates that \hspg{} reaches comparable convergence as~\proxsg{} and~\proxsvrg{} in practice.  Finally, Figure~\ref{figure:runtime_convex} indicates that Prox-SG, RDA and \hspg{} have similar computational cost to proceed, except~\proxsvrg{} due to its periodical full gradient computation.

\vspace{-0.2cm}
\begin{table}[h]
	\centering
	\def\arraystretch{1.1}
	\caption{Summary of datasets.\label{table:datasets}}
	\resizebox{\textwidth}{!}{
		\begin{tabular}{@{\extracolsep{4pt}}ccccccccc}
			\Xhline{2\arrayrulewidth}
			Dataset & N & n  & Attribute & & Dataset & N & n  & Attribute \\
			\hline
			a9a & 32561 & 123 & binary \{0, 1\} & & news20 & 19996 & 1355191 &   unit-length \\
			higgs & 11000000 & 28 & real $[-3, 41]$ & & real-sim & 72309 & 20958 & real [0, 1]\\
			ijcnn1 & 49990 & 22  &  real [-1, 1] & &  url\_combined & 2396130 & 3231961 & real $[-4, 9]$ \\
			kdda & 8407752 & 20216830 & real $[-1, 4]$ & & w8a & 49749 & 300   & binary \{0, 1\}\\
			\Xhline{2\arrayrulewidth}
		\end{tabular}
	}
\end{table}
\vspace{-0.2cm}
\begin{table}[h]
	
	\centering
	\def\arraystretch{1.1}
	
	\caption{Final objective values $\psi$ for tested algorithms on convex problems.}
	\label{table:object_Psi_value_convex}
	{\scriptsize
		\begin{tabularx}{\textwidth} { 
				>{\centering\arraybackslash}X 
				>{\centering\arraybackslash}X 
				>{\centering\arraybackslash}X 
				>{\centering\arraybackslash}X 
				>{\centering\arraybackslash}X 
				>{\centering\arraybackslash}X  }
			\Xhline{3\arrayrulewidth}
			\multirow{2}{*}{Dataset} & \multirow{2}{*}{\proxsg{}} & \multirow{2}{*}{\rda} & \multirow{2}{*}{\proxsvrg{}}  & \multicolumn{2}{c}{\hspg{}} \\
			\cline{5-6}
			& &  & &  $\epsilon$ as $0$ & $\epsilon$ as $0.8$\\
			\hline
			a9a & \textbf{0.355} & 0.359  & \textbf{0.355} & \textbf{0.355} & \textbf{0.355} \\
			higgs & \textbf{0.357} & 0.360 & 0.365 & 0.358 & 0.358\\
			ijcnn1 & \textbf{0.248} & 0.278 & \textbf{0.248} & \textbf{0.248} & \textbf{0.248}\\
			kdda & \textbf{0.103} & 0.124 & \textbf{0.103} & \textbf{0.103} & \textbf{0.103}\\
			news20 & \textbf{0.538} & 0.693 & \textbf{0.538} & \textbf{0.538} & \textbf{0.538}  \\
			real-sim & \textbf{0.242} & 0.666 & 0.244 & \textbf{0.242} & \textbf{0.242} \\
			url\_combined & 0.397 & 0.579  & \textbf{0.391} &  0.405 & 0.405\\
			w8a & \textbf{0.110} & 0.111 & 0.112 & \textbf{0.110} & \textbf{0.110}\\
			\Xhline{3\arrayrulewidth} 
		\end{tabularx}
	}
	\caption{Group sparsity for tested algorithms on convex problems.}
	\label{table:group_sparsity_convex}
	
	{\scriptsize
		\begin{tabularx}{\textwidth} { 
				>{\centering\arraybackslash}X 
				>{\centering\arraybackslash}X 
				>{\centering\arraybackslash}X 
				>{\centering\arraybackslash}X 
				>{\centering\arraybackslash}X 
				>{\centering\arraybackslash}X }
			\Xhline{3\arrayrulewidth}
			\multirow{2}{*}{Dataset} & \multirow{2}{*}{\proxsg{}} & \multirow{2}{*}{\rda{}} & \multirow{2}{*}{\proxsvrg{}} & \multicolumn{2}{c}{\hspg{}} \\
			\cline{5-6}
			& &  & &  $\epsilon$ as $0$ & $\epsilon$ as $0.8$\\
			\hline
			a9a & 20\% & \textbf{30\%}  & \textbf{30\%} & \textbf{30\%} & \textbf{30\%} \\
			higgs & 0\% & 10\% & 0\% & 0\% & \textbf{30\%}\\
			ijcnn1 & 50\% & \textbf{70\%} & 60\% & 60\% & 60\% \\
			kdda & 0\% & 0\% & 0\% & 0\% & \textbf{80\%}\\
			news20 & 20\% & 80\% & \textbf{90\%} & 80\% & \textbf{90\%}  \\
			real-sim & 0\% & 0\% & \textbf{80\%} & 0\% & \textbf{80\%} \\
			url\_combined & 0\% & 0\% & 0\% & 0\% & \textbf{90\%} \\
			w8a & \textbf{0\%} & \textbf{0\%} & \textbf{0\%} & \textbf{0\%} & \textbf{0\%} \\
			\Xhline{3\arrayrulewidth} 
		\end{tabularx}
	}
\end{table} 

\subsection{Nonconvex Experiments}\label{appendix.hspg.experiment.nonconvex}
To illustrate, among the state-of-the-art proximal stochastic optimizers, we exclude~\rda{} because of no acceptable results attained during our following tests with the step size parameter $\gamma$ setting throughout all powers of 10 from $10^{-3}$ to $10^3$, and skip Prox-Spider and SAGA since Prox-SVRG has been a superb representative to the proximal incremental gradient methods. We consider the popular image classification tasks, with popular architectures, \ie,  \vgg{} and \resnet{} on benchmark datasets~\cifar{} and~\fashionmnist{}~\cite{xiao2017online}, where the group partition $\mathcal{G}$ is defined as 3D kernel following~\cite{deleu2021structured,lin2019toward}, which are not ZIGs.

\begin{table*}[ht]
	\centering
	\caption{Final $\psi$/group sparsity ratio/testing accuracy on non-convex problems over non-ZIGs.}
	\label{table:nonconvex}
	\resizebox{\textwidth}{!}{
		\begin{tabular}{ 
				cccccc}
			\Xhline{3\arrayrulewidth}
			Backbone & Dataset & \proxsg{} & \proxsvrg{}  & \hspg{}\\
			\hline
			\multirow{2}{*}{\vgg{}} & \cifar{} & 0.59\ /\ 52.58\%\ /\ 90.50\%  & 0.85\ /\ 14.13\%\ /\ 89.16\% & \textbf{0.58}\ /\ \textbf{76.47\%} \ /\ \textbf{91.93\%}  \\ 
			& \fashionmnist{} & 0.52\ /\ 12.31\%\ /\ \textbf{92.83\%}  & 2.66\ /\ 0.38\%\ /\ 92.72\% &  \textbf{0.52}\ /\ \textbf{47.82}\%\ /\ 92.87\% \\\hdashline 
			\multirow{2}{*}{\resnet{}} & \cifar{} & \textbf{0.31}\ /\ 20.27\%\ /\ 94.36\%  & 0.37\ /\ 4.60\%\ /\ 94.11\%  & \textbf{0.31}\ /\ \textbf{69.98\%}\ /\ \textbf{94.40\%} \\ 
			& \fashionmnist{} & 0.14\ /\ 0.00\%\ /\ \textbf{94.94}\% & 0.18\ /\ 0.00\%\ /\ 94.70\% & \textbf{0.13}\ /\ \textbf{77.08\%}\ /\ 94.61\%\\ 
			\hdashline
			\multirow{2}{*}{\mobilenet{}} & \cifar{} & \textbf{0.40}\ /\ 58.05\% \ /\ 91.54\% & 0.65\ /\ 29.20\% \ /\ 89.68\% & \textbf{0.40}\ /\ \textbf{71.36\%} \ /\ 92.04\% \\ 
			& \fashionmnist{} & \textbf{0.22}\ /\ 62.62\% \ /\ 94.22\%  & 0.40\ /\ 41.99\%  \ /\ 94.19\% & 0.26\ /\ \textbf{84.26\%}\ /\ \textbf{94.52\%} \\ 
			\Xhline{3\arrayrulewidth} 
	\end{tabular}}
\end{table*} 

Table~\ref{table:nonconvex} demonstrates the effectiveness and superiority of~\hspg{}, where we mark the best values as bold, and the group sparsity ratio is defined as the percentage of zero groups. In particular, \textit{(i)} \hspg{} computes remarkably higher group sparsity than other methods on all tests, of which the solutions are typically multiple times sparser in the manner of group than those of~\proxsg{}, while~\proxsvrg{} performs not comparably since the variance reduction techniques may not work as desired for deep learning applications; \textit{(ii)} \hspg{} performs competitively with respect to the final objective values $\psi$. In addition, all the methods reach a comparable generalization performance on unseen test data. On the other hand, sparse regularization methods may yield solutions with entries that are not exactly zero but are very small. Sometimes all entries below certain threshold ($\mathcal{T}$) are set to zero~\cite{el2018combinatorial}. However, such simple truncation mechanism is heuristic-rule based, hence may hurt convergence and accuracy. To illustrate this, we set the groups of the solutions of~\proxsg{} and~\proxsvrg{} to zero if the magnitudes of the group variables are less than some $\mathcal{T}$, and denote the corresponding solutions as \proxsg{}* and~\proxsvrg{}*. 

\begin{figure}[h]
	\centering
	\begin{subfigure}[t]{0.4\textwidth}
		\includegraphics[width=\linewidth]{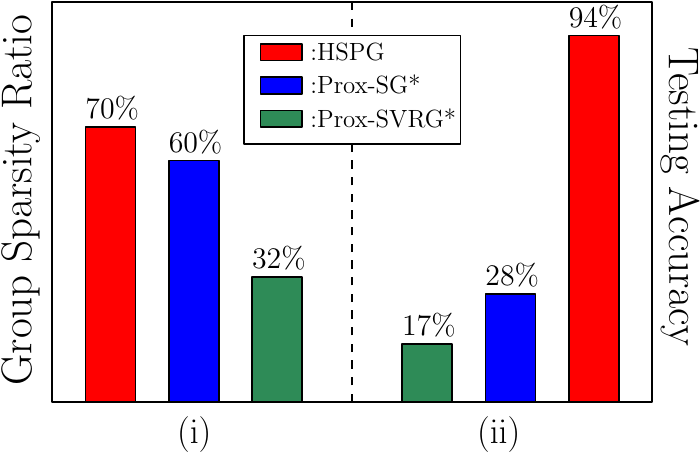}
		\caption{HSPG VS Truncation over non-ZIGs.}
		\label{figure:simple_truncation_non_zigs}
	\end{subfigure}
	\hspace{0.2mm}
	\begin{subfigure}[t]{0.4\textwidth}
		\includegraphics[width=\linewidth]{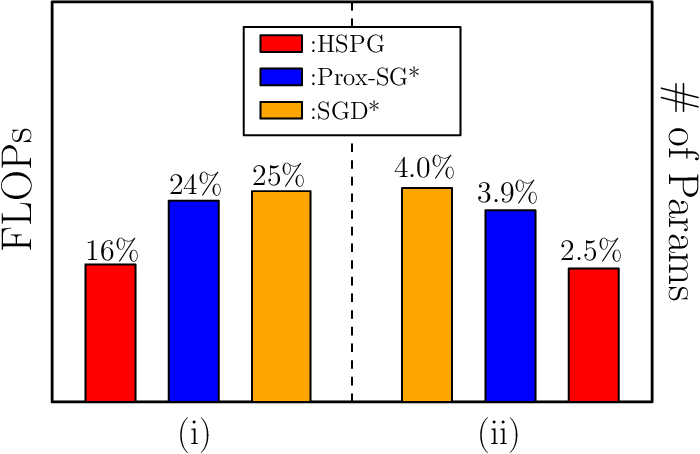}
		\caption{HSPG VS Truncation over ZIGs.}
		\label{figure:simple_truncation_zigs}
	\end{subfigure}
	\caption{\small HSPG versus simple truncation. (a) On~\resnet{} with~\cifar{} over non-ZIGs. (b) On~\vgg{} with~\cifar{} over ZIGs.}
\end{figure}

As shown in Figure~\ref{figure:simple_truncation_non_zigs}\textcolor{red}{(i)}, under the $\mathcal{T}$ with no accuracy regression,~\proxsg{}* and~\proxsvrg{}* reach higher group sparsity ratio as 60\% and 32\% compared to Table~\ref{table:nonconvex}, but still significantly lower than the 70\% of HSPG without simple truncation. Under the $\mathcal{T}$ to reach the same group sparsity ratio as HSPG, the testing accuracy of~\proxsg{}* and~\proxsvrg{}* regresses drastically to 28\% and 17\% in Figure~\ref{figure:simple_truncation_non_zigs}\textcolor{red}{(ii)} respectively. Remark here that although further refitting the models from \proxsg{}* and~\proxsvrg{}* on active (non-zero) groups of weights may recover the accuracy regression, it requires additional engineering efforts and training cost, which is less attractive and convenient than~\hspg{} (with no need to refit). Similarly, as shown in Figure~\ref{figure:simple_truncation_zigs}, under the ZIG partition and the $\mathcal{T}$ without accuracy regression, the FLOPs and number of parameters reductions achieved by SGD* (subgradient descent with simple truncation) and~\proxsg{}* are not comparable with those achieve by~\hspg{},~\ie,~\hspg{} achieves about $1.5\times$ fewer FLOPs and number of parameters.

\end{document}